\documentclass[11pt, a4paper, twocolumn, copyright, goog]{google}

\usepackage[authoryear, compress, round]{natbib} 
\usepackage{multirow}
\usepackage{csquotes}
\usepackage{comment}
\usepackage{svg}
\definecolor{ncarow}{rgb}{0.94, 0.93, 0.98}
\usepackage{float}


\uselogo{} 

\title{Reasoning with Neural Cellular Automata}

\correspondingauthor{mayalen@google.com}

\author[1]{Mayalen Etcheverry}
\author[1]{Pietro Miotti}
\author[1,2,3]{Aidan Sirbu}
\author[1]{Konstantin Schürholt}
\author[1,4]{Mariia Drozdova}
\author[1]{Arna Ghosh}
\author[1]{Blaise Agüera y Arcas}
\author[1]{James Manyika}
\author[*,1,2,3,5,6,7]{Blake Richards}
\author[*,1]{Eyvind Niklasson}

\affil[1]{Google Paradigms of Intelligence Team}
\affil[2]{School of Computer Science, McGill University}
\affil[3]{Mila - Quebec AI Institute}
\affil[4]{University of Geneva}
\affil[5]{Department of Neurology and Neurosurgery, McGill University}
\affil[6]{Montreal Neurological Institute, McGill University}
\affil[7]{Learning in Machines and Brains Program, CIFAR}
\affil[*]{Equal supervision}

\begin{abstract}
Modern AI architectures used to solve visual reasoning tasks typically rely heavily on global connectivity and synchronization. As biological systems demonstrate, though, sophisticated computation can be performed in a more decentralized fashion. In this work, we test the reasoning capabilities of Neural Cellular Automata (NCAs), networks of recurrent cells that use strictly local connectivity and asynchronous updates. NCAs have been extensively studied in artificial life experiments, but it is unclear whether they can perform complex multi-step reasoning. We show that NCAs produce spatio-temporal dynamics capable of solving challenging visual reasoning tasks, including large mazes, Sudoku, and ARC-AGI-1. Furthermore, we provide evidence that NCAs generalize out-of-distribution when running with larger grids, longer rollouts, or parallel trials; and that the latter can be made more efficient via pruning of redundant trajectories. We find that these generalization capabilities depend on training with sample replay and stochastic perturbations, and that stochasticity remains beneficial at test time. Finally, we show that NCAs are robust reasoners capable of dynamically modulating compute to recover efficiently from damage, and that they can scale to solve reasoning in raw pixel space.
\end{abstract}

\begin{document}

\maketitle

\section{Introduction}

Self-organization, the process by which low-level units interact locally to produce sophisticated global patterns, is a staple of biological life and intelligence~\citep{Camazine2003,Karsenti2008}.
The degree of self-organization in a computational system can range across a spectrum, from purely centralized architectures to highly localized, decentralized interactions. Most modern artificial intelligence (AI) systems use distributed representations, but they still rely heavily on global connectivity and synchronized interaction between units, placing them farther along the centralization spectrum.  Moreover, while architectures such as transformers and their looped variants have been widely used in recent years, their all-to-all connectivity and associated data-movement dictate energy cost, and fundamentally set limits on their underlying physical computing paradigms. Enforcing stronger locality constraints from the ground up could offer a principled alternative for AI models that run on decentralized, fully self-organizing computing substrates. 
However, it remains an open question to what extent locality constrained architectures can actually be scaled, and whether they can really perform complex tasks to address modern AI challenges. In particular, it remains unclear whether low-level, localized communication protocols can solve complex reasoning tasks. Reasoning is a cornerstone of general intelligence that presents a notorious challenge even for modern deep learning architectures. Classical tasks such as pathfinding (e.g., mazes) and constraint satisfaction problems (e.g., Sudoku), along with recent benchmarks like ARC-AGI, are being extensively used to benchmark the ability of AI systems to solve logical problems and generalize out-of-distribution. Solving these tasks requires AI models to jointly learn to infer the underlying logic and to execute the multi-step procedures necessary to generate valid solutions, all while being trained solely on input-output observations. Current research typically follows one of two approaches: prompting large, generalist LLMs with in-context task information; or using recursive reasoning approaches with compact specialist models such as HRM~\citep{Wang2025}, TRM~\citep{Jolicoeur-Martineau2025} or LoopViT~\citep{Shu2026}.
Yet, most reasoning models developed to date still rely heavily on dense, longe-range connectivity and synchronized execution, allowing every part of the context to be attended to at each update step.

Here we explore whether there is truly a need for global connectivity and coordination in order to engage in effective reasoning. Specifically, we investigate whether Neural Cellular Automata (NCA), a recent neural network architecture that uses fully local connectivity between a distributed grid of recurrent cells~\citep{Mordvintsev2020}, can support distributed reasoning. At the intersection of Artificial Life and AI, and inspired by research on self-organization and morphogenesis, NCAs can be viewed as specific instances of recurrent convolutional or graph neural networks with a strong locality constraint. They are also closely related to recent recursive reasoning architectures, with the central difference being their use of \textit{local connectivity} and \textit{asynchronous updates}. NCAs have already been applied across diverse domains, from pattern generation and repair~\citep{Mordvintsev2020, Kim2026} to classification~\citep{Randazzo2020}, segmentation~\citep{Sandler2020}, control~\citep{Variengien2021}, and as ViT adaptor layers~\citep{Xu2024}, showing compelling properties such as high parameter efficiency and robustness. Although recent studies demonstrated initial success applying NCAs to mazes~\citep{Earle2023} and ARC-AGI-1~\citep{Etienne2025}, their scaling behavior and potential for reasoning remain largely unexplored.

In this work, we demonstrate the reasoning capabilities of NCAs across a suite of reasoning benchmarks: pathfinding on large out-of-distribution mazes~\citep{Schwarzschild2021} and multi-solution ones~\citep{Wang2025}; constraint-satisfaction on out-of-distribution~\citep{Miyato2024} and extremely difficult~\citep{Wang2025} Sudoku boards; program induction and few-shot visual reasoning on the ARC-AGI-1 benchmark~\citep{Chollet2019}; and joint perception-reasoning directly in pixel space on Visual Sudoku~\citep{Wang2019}. 
Our main contributions are fourfold.
First, we show that, despite their strong locality constraints, NCAs perform well on all these benchmarks using compact architectures. With few parameters, cells locally coordinate to solve complex multi-step reasoning tasks, with the spatial communication bottleneck giving rise to structured, observable reasoning dynamics that unfold across both space and time. On ARC-AGI-1, we show that a single NCA can solve multiple tasks and transfer across them when provided with appropriate contextual information. 
Second, we demonstrate that NCAs can generalize to harder, out-of-distribution tasks when compute is expanded at test-time across either space or time, or via parallelization. To mitigate the cost of running multiple, parallel rollouts, we propose a pruning strategy that shows more efficient scaling for low compute budgets. We show that the generalization capabilities of these recursive, distributed reasoners depend on training with sample replay and stochastic perturbations, and that stochasticity remains beneficial at test time. 
Third, we highlight compelling properties of NCAs, revealing that NCAs are robust and adaptive reasoners that can dynamically modulate their compute to solve tasks and repair from damage more efficiently. 
Finally, we show that NCAs can reason in larger, harder problem modalities and solve Sudokus directly in pixel space.

\begin{figure*}
  \centering
  \includegraphics[width=1.0\textwidth]{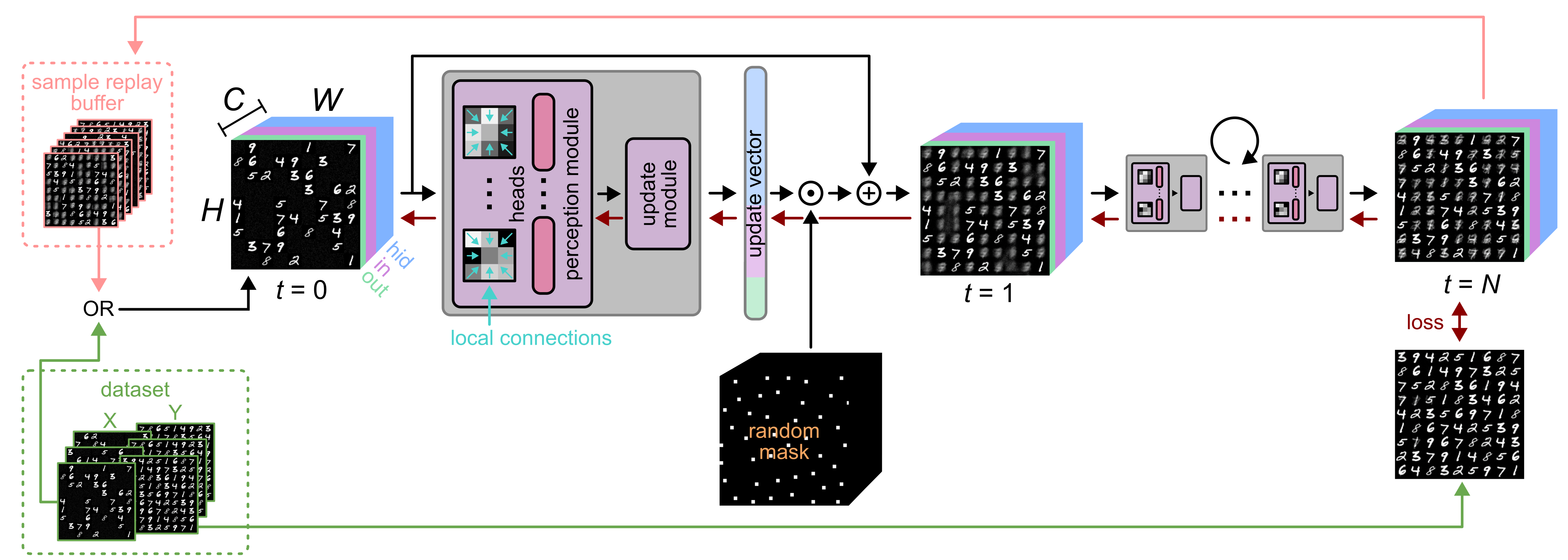}  
  \caption{\textbf{Overview of reasoning with Neural Cellular Automata (NCA).} Cells maintain structured states ($C_{\text{in}}$, $C_{\text{out}}$, $C_{\text{hid}}$) on an $H \times W$ grid and update iteratively using local $3 \times 3$ perception and a shared update module, applied residually under a stochastic firing mask. Training unrolls grids from the dataset or a sample replay buffer for $N$ steps, optimizing the NCA to solve the task.}
  \label{figure_arch_diagram}
\end{figure*}

\section{Related Work: Recurrent Visual Reasoning (RvR)}

\paragraph*{RvR Architectures.} Most recent RvR models use variants of looped transformers with global attention~\citep{Miyato2024,Wang2025,Jolicoeur-Martineau2025,Baek2026}. LoopViT uses a hybrid architecture that alternates global and local attention layers~\citep{Shu2026}. Some prior works use fully-local NCAs for pathfinding~\citep{EndoOther,Earle2023} with some degree of length generalization using handcoding-inspired variants, and for ARC-AGI-1 achieving 12.9\% on a subset of 262 fixed-size tasks by training one NCA per task~\citep{Etienne2025}. Sheaf-ADMM~\citep{Seely2026} uses a NCA-like architecture with local agents for visual reasoning, but relies on non-strict locality for Sudoku (i.e. agents can see full rows, columns or blocks) and shows very limited length-generalization on mazes. In our work, cells coordinate under strictly local observability to solve complex problems including Sudoku-Extreme and multi-task ARC.  

\paragraph*{RvR Training.} Training NCAs with \textit{sample replay}~\citep{Mordvintsev2020,Du2019} parallels strategies like progressive loss~\citep{Bansal2022} and deep supervision~\citep{Jolicoeur-Martineau2025} used in the recent RvR literature to emulate extended rollouts and encourage convergence to stable fixed-points without deep unrolling. Exploiting \textit{stochasticity} is also proposed by other recent RvR work~\citep{Baek2026}, but they do so by injecting some noise in part of the state used for prediction, whereas we use asynchronous updates and perturbations across the full state. Other concurrent works~\citep{Helbling2026,Drozdova2026,Suleymanzade2026} propose that diffusion-inspired progressive noise and local denoising objectives may provide another effective alternative for regularizing the convergence landscape of RvR models.

\paragraph*{RvR Test-Time Scaling.} RvR models are dynamical systems with complex state spaces~\citep{Lai2026}. Several works, including ours, leverage stochastic parallel trials for exploring the state-space at test-time, effectively increasing the chance of discovering the correct solution with some variants in the candidate selection mechanism~\citep{Miyato2024,Sghaier2026,Baek2026}. In addition, we propose \textit{Niche-Capped Diversity Pruning} to discard redundant trajectories and optimize test-time scaling for constrained compute budgets (see section \ref{subsec_ood_generalization}).

\paragraph*{RvR with Adaptive Compute.} Several RvR models use global convergence metrics as early-exit mechanisms, such as learned halting~\citep{Jolicoeur-Martineau2025} or entropy monitoring~\citep{Shu2026}. While these methods act as global on/off switches that stop all compute at once, our approach decentralizes adaptive compute: cells modulate their update frequency independently based on self-predicted confidence. This creates an organic allocation of compute that naturally concentrates on unresolved or perturbed sub-regions, while stable cells idle and save resources (see section \ref{subsec_robust_adaptive_reasoners}).

\section{Method}
\label{sec_method}

In this section, we introduce the NCA architecture (Figure~\ref{figure_arch_diagram}) and the associated training regimes and test-time scaling procedures that we use for recursive reasoning. Complete architectural, training, and testing details are provided in Appendix~\ref{sec_method_details}, and hyperparameters in Appendix~\ref{sec_experimental_settings}.

\subsection{Architecture}

\paragraph*{NCA Cell State.} The grid state is represented as $X \in \mathbb{R}^{H \times W \times C}$, where spatial dimensions $H \times W$ mirror the problem topology (e.g., $9 \times 9$ for Sudoku) and each cell maintains a state vector $x_{i,j} \in \mathbb{R}^C$ across $C$ \textit{channels}. The state is partitioned into functional slices: output slice ($C_{\text{out}}$) for token predictions, hidden slice ($C_{\text{hid}}$) for latent reasoning, an immutable input slice ($C_{\text{in}}$) for task constraints, and an additional task slice ($C_{\text{task}}$) for ARC-AGI-1. Unlike prior NCA works operating directly on RGB pixel values, our tasks use discrete vocabularies (e.g. digits for Sudoku and color symbols for ARC) mapped to fixed orthogonal embedding vectors that are loaded directly into $C_{\text{in}}$ at initialization ($t=0$). Depending on the task, the input $C_{\text{in}}$ slice acts as environmental constraints (e.g., spanning all channels in mazes to block information flow through walls) or as environmental clues (e.g., overlapping with $C_{\text{out}}$ in Sudoku to clamp known clues on the prediction channels, or forming a dedicated read-only slice in ARC-AGI-1). For ARC, the mutable $C_{\text{task}}$ slice is initialized at $t=0$ with a learned global task embedding to condition cells on the target task.

\paragraph*{NCA Cell Update.}
At each step, cells update asynchronously in two stages: (i)~\textit{Perception}, where a multi-head perception module maps immediate $3 \times 3$ neighbor states (Moore neighborhood) into a vector $z_{i,j}$; and (ii)~\textit{Update}, where a shared, weight-tied MLP maps the perception vector $z_{i,j}$ to an update $\Delta x_{i,j}$, applied residually with a stochastic update mask $m_{i,j} \sim \text{Bernoulli}(p_{\text{fire}})$: $x_{i,j}^{(t+1)} = x_{i,j}^{(t)} + m_{i,j} \Delta x_{i,j}$. For perception, we use learned convolutions for Maze and Visual Sudoku, and a position-dependent variant of attention called \enquote{fixed attention} for Sudoku and ARC. While standard self-attention is more expressive, we observed that attention weights converged to depend purely on relative positions rather than cell states (Figure~S\ref{figure_sa_convergence}); replacing it with fixed attention improved performance (Figure~S\ref{figure_sa_vs_fa}).
After the update, we apply local-only normalization to cells for Sudoku and ARC: channels are split into groups and normalized to unit length similar to \cite{Miyato2024}.

\paragraph*{NCA Cell Prediction.}
At any step, the output slice $x_{i,j,\text{out}}$ can be mapped to logit predictions via cosine similarity to token embeddings (Maze, Sudoku) or using a lightweight MLP head (ARC). 
A per-cell confidence $c_{i,j} \in [0, 1]$ is derived from the output distribution (e.g., maximum probability).

\subsection{Training}

\paragraph*{Training with Sample Replay.} Following the original NCA training pipeline~\citep{Mordvintsev2020}, models are trained using Backpropagation Through Time (BPTT) coupled with a \textit{sample replay buffer} and \textit{training perturbations}. Training batches mix previously evolved grid states drawn from the buffer with fresh initializations from the dataset, with stochastic perturbations (Gaussian noise, damage, and target swap). The batch is then unrolled under the NCA update rule for $N$ steps to compute gradients via BPTT, and post-rollout states are written back into the buffer in place. 

\paragraph*{Training with Time Encodings (ARC). } For ARC, we adapt the pre-training and test-time-training (TTT) pipeline from recent VARC~\citep{Hu2025} and LoopViT~\citep{Shu2026} works. Models are trained via BPTT for $N=64$ steps, using the RE-ARC dataset and online augmentations (reflections, color permutations, translations). A learned time embedding is added to the hidden state of every cell at every step. At inference, we replace this global step with individual cell clocks: each cell maintains an internal, locally-incremented counter that advances only when the cell fires.

\subsection{Test-time Scaling}

\paragraph*{Parallel Trials Scaling.} NCA rollouts are inherently stochastic due to asynchronous updates and random initializations. Consequently, running $K$ parallel rollouts for the same input task produces distinct trajectories that increase the chance of converging to a valid solution (parallel trials). We then select the final prediction among the $K$ candidates via either highest board-level confidence (average of all per-cell confidences $c_{i,j}$ for Maze and Sudoku) or via majority voting (ARC). 

\paragraph*{Niche-Capped Diversity Pruning.} To encourage parallel compute to be allocated across structurally diverse solution hypotheses rather than wasted on duplicate paths, we introduce \textit{Niche-Capped Diversity Pruning} for parallel trials scaling. At scheduled rollout checkpoints, active trajectories are grouped into discrete solution \enquote{niches} based on their current predictions; and the ensemble size is progressively halved by capping each niche at a maximum capacity (see Appendix~\ref{subsec_tts_pruning_details}).

\paragraph*{Spatial Substrate Scaling.}
Because the cells that compose NCAs rely strictly on local perception they are agnostic to grid size; the computational substrate can be expanded at test time simply by tiling additional cells. We exploit this form of scaling for Maze-OOD, where cell perception is position-independent and the grid can be seamlessly expanded with no weight updates.

\section{Experimental Results}
\subsection{Experimental Setup}

\paragraph*{Benchmarks.} We evaluate NCAs across various spatial reasoning tasks detailed in Appendix~\ref{sec_benchmarks}: (i)~\textit{Maze path-finding}, testing out-of-distribution size generalization (\textbf{Maze-OOD}; \citealp{Bansal2022}) and solution ambiguity on hard mazes (\textbf{Maze-Hard}; \citealp{Wang2025}); (ii)~\textit{Sudoku distributed constraint satisfaction problem (DiSCP)}, testing out-of-distribution difficulty generalization (\textbf{Sudoku-OOD}; \citealp{Miyato2024}) and challenging search and backtracks on very hard boards (\textbf{Sudoku-Extreme}; \citealp{Wang2025}); (iii)~\textit{Few-shot abstraction}, testing task-conditioned program induction from few input-output demonstrations and augmentations (\textbf{ARC-AGI-1}; \citealp{Chollet2019}); and (iv)~\textit{Pixel-space reasoning}, testing end-to-end perception and reasoning on much larger raw image inputs (\textbf{Visual-Sudoku}).

\paragraph*{Evaluation.} For each benchmark, we train a NCA using the architecture and training pipelines detailed in Appendix~\ref{sec_method_details}, and experimental settings provided in Appendix~\ref{sec_experimental_settings}. Inference runs $K$ parallel rollouts of the trained model, each for $D$ iterations. The values of $D, K$, and grid size ($S=H \times W$) used for each benchmark are specified in Table~\ref{table_results_across_benchmarks}.

\paragraph*{Baselines.} Our objective is not to establish new state-of-the-art results, but to demonstrate that NCAs can execute complex reasoning tasks in a strictly local, fully distributed fashion. 
Nonetheless, we include top-performing RvR baselines as reference points to contextualize NCA results: DeepThink \citep{Bansal2022} for Maze-OOD;
PTRM \citep{Sghaier2026} for Maze-Hard;  
AKOrN \citep{Miyato2024} for Sudoku-OOD;
PTRM for Sudoku-Extreme; and
TRM and LoopViT~\citep{Jolicoeur-Martineau2025,Shu2026} for ARC-AGI-1. 

\begin{table}[h!]
  \centering
  \setlength{\tabcolsep}{2pt} 
  
  \caption{Overview of NCA performances and efficiency across reasoning benchmarks (see section~\ref{subsec_flops_estimation} for FLOPs estimation).}
  \label{table_results_across_benchmarks}

  \footnotesize
  
   \begin{tabular}{lccc}
    \toprule
    Model \scriptsize{{(test config)}} & Params & FLOPs & Acc (\%) \\
    \hline
    \multicolumn{4}{l}{\cellcolor{gray!10}{\textit{Maze-OOD} \scriptsize{(K=1)}}} \\
    \hline
    DeepThink \scriptsize{(S=13x13, D=100)} & 784K & 160.5B & 99.9 \\
    \rowcolor{ncarow} \textbf{NCA} \scriptsize{(S=13x13, D=300)} & 10K & 1.5B & 100.0 \\
    DeepThink \scriptsize{(S=59x59, D=1000)} & 784K & 24.1T & 97.3 \\
    \rowcolor{ncarow} \textbf{NCA} \scriptsize{(S=59x59, D=2000)} & 10K & 163.4B & 100.0 \\
    DeepThink \scriptsize{(S=201x201, D=2000)} & 784K & 521.7T & 74.0 \\
    \rowcolor{ncarow} \textbf{NCA} \scriptsize{(S=201x201, D=13000)} & 10K & 11.8T & 100.0 \\
    \hline
    \multicolumn{4}{l}{\cellcolor{gray!10}{\textit{Maze-Hard} \scriptsize{(S=30x30)}}} \\
    \hline
    PTRM \scriptsize{(D=16, K=1)} on A* & 7M & 3.8T & 83.8 \\
    \rowcolor{ncarow} \textbf{NCA} \scriptsize{(D=200, K=1)} on A*/BFS & 41K & 15.3B & 86.6/94.7 \\
    PTRM \scriptsize{(D=16, mode@K=100)} & 7M & 382.2T & 86.7 \\
    \rowcolor{ncarow} \textbf{NCA} \scriptsize{(D=200, conf@K=128)} & 41K & 2.0T & 89.2/96.2  \\
    \hline
    \multicolumn{4}{l}{\cellcolor{gray!10}{\textit{Sudoku-OOD} \scriptsize{(S=9x9)}}} \\
    \hline
    AKOrN \scriptsize{(D=128, K=1)} & 3M & 23.8B & 51.7 \\
    \rowcolor{ncarow} \textbf{NCA} \scriptsize{(D=256, K=1)} & 160K & 11.7B & 66.1 \\
    AKOrN \scriptsize{(D=128, conf@K=4096)} & 3M & 97.3T & 89.5 \\
    \rowcolor{ncarow} \textbf{NCA} \scriptsize{(D=256, conf@K=2048)} & 160K & 23.9T & 98.5 \\
    \hline
    \multicolumn{4}{l}{\cellcolor{gray!10}{\textit{Sudoku-Extreme} \scriptsize{(S=9x9)}}} \\
    \hline
    PTRM \scriptsize{(D=64, K=1)} & 5M & 1.6T & 87.3 \\
    \rowcolor{ncarow} \textbf{NCA} \scriptsize{(D=2048, K=1)} & 603K & 371.3B & 75.6 \\
    PTRM \scriptsize{(D=64, conf@K=100)} & 5M & 164.0T & 98.8 \\
    \rowcolor{ncarow} \textbf{NCA} \scriptsize{(D=2048,  conf@K=512)} & 603K & 190.1T & 92.7 \\
    
    \hline
    \multicolumn{4}{l}{\cellcolor{gray!10}{\textit{ARC-AGI-1}}} \\
    \hline
        TRM \scriptsize{(D=16, pass@2, K=1000)} &   7M & 3.8P &  44.6  \\
        \rowcolor{ncarow} \textbf{NCA} \scriptsize{(D=64, pass@2, K=3264)} &   3.5M & 2.4P & 48.8 \\
        \rowcolor{ncarow} \textbf{NCA} \scriptsize{(D=64, pass@64, K=3264)} &  3.5M & 2.4P & 60.3 \\
        \rowcolor{ncarow} \textbf{NCA E} \scriptsize{(D=64, pass@2, K=9792)}&   10.5M &  7.1P & 51.3 \\
        \rowcolor{ncarow} \textbf{NCA E} \scriptsize{(D=64, pass@64, K=9792)} &  10.5M & 7.1P & 63.0 \\
        LoopVIT s \scriptsize{(D=24, pass@2, K=510)} & 3.8M & 80.1T & 60.1  \\
        LoopVIT m \scriptsize{(D=8, pass@2, K=510)} & 11.3M & 132.7T & 63.8  \\
        LoopVIT l \scriptsize{(D=8, pass@2, K=510)} & 17.6M & 220.5T & 65.8  \\
    \hline
    \multicolumn{4}{l}{\cellcolor{gray!10}{\textit{Visual-Sudoku} \scriptsize{(S=256x256, K=1)}}} \\
    \hline
        \rowcolor{ncarow} \textbf{NCA} \scriptsize{(D=1024) on easy (ID)} & 2.1M & 283.2T & 87.8 \\
        \rowcolor{ncarow} \textbf{NCA} \scriptsize{(D=3072) on hard (OOD)} & 2.1M & 849.5T & 18.9 \\
    \bottomrule
    \end{tabular}
\end{table}

\begin{figure*}
    \definecolor{mgray}{RGB}{128, 128, 128}
    \definecolor{myellow}{RGB}{240, 240, 0}
    \definecolor{mpurple}{RGB}{230, 0, 230}
    \definecolor{morange}{RGB}{255, 140, 0}
    \definecolor{mblue}{RGB}{30, 144, 255}
    \definecolor{mgreen}{RGB}{60, 179, 113}
    \definecolor{sred}{RGB}{200, 0, 0}
    \definecolor{sblue}{RGB}{0, 0, 200}
    \definecolor{sgreen}{RGB}{0, 120, 0}
  \centering
  \includegraphics[width=\textwidth]{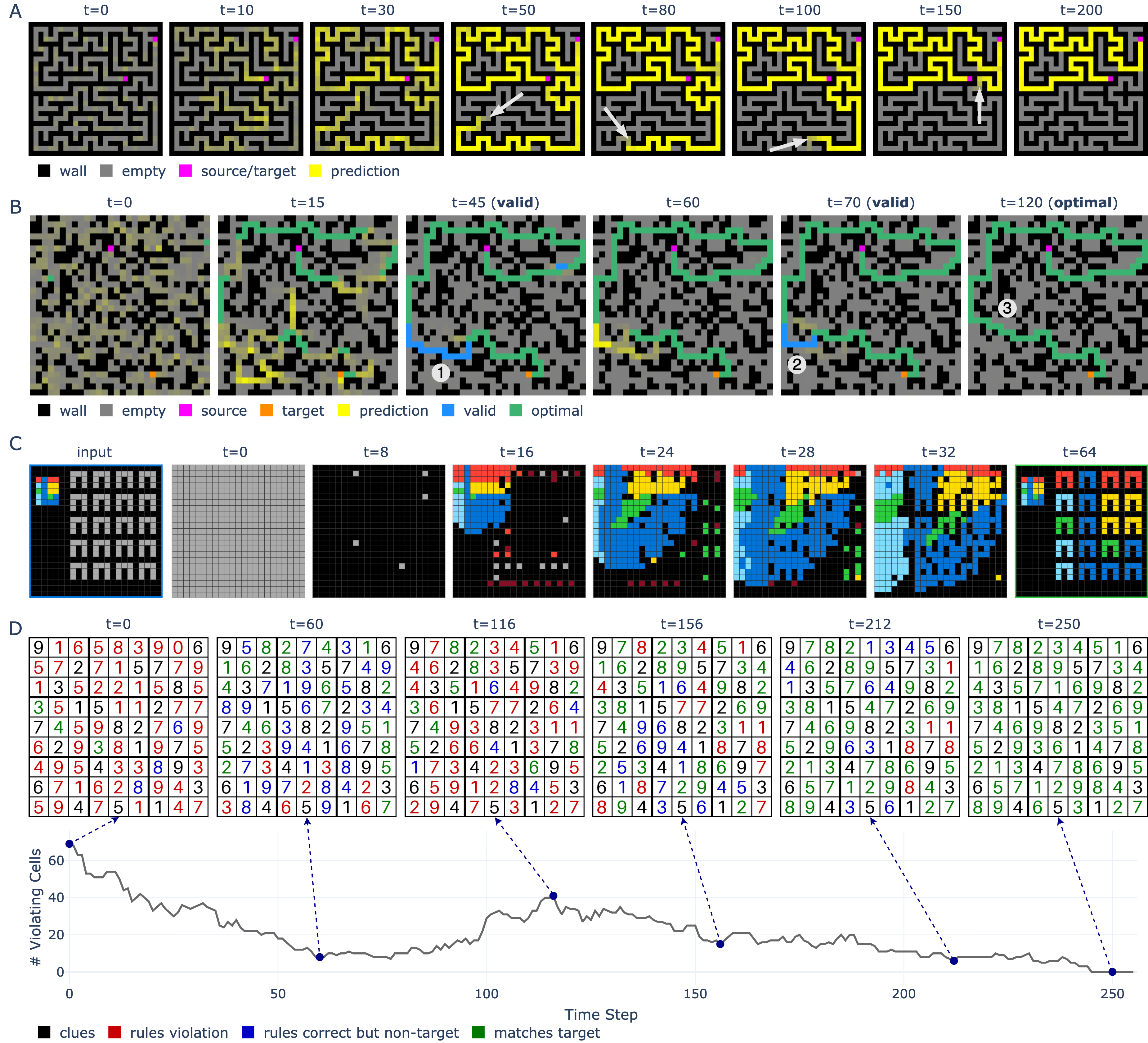}
  \caption{Emergent reasoning dynamics in NCAs.
  \textbf{(A) Backtracking in mazes:} cell activations explore paths in parallel, then locally back-propagate \enquote{waves} to prune dead-end paths, until converging to the final solution path.
  \textbf{(B) Iterative trajectory refinement in Maze-Hard:} cells rapidly resolve unambiguous segments ($t=15$) until settling first on a valid, sub-optimal solution (1, blue), before discovering a shorter valid alternative (2), and ultimately stabilizing into an optimal solution (3, green).
  \textbf{(C) Local spatial propagation of objects and colors in ARC-AGI-1:} solving the task of coloring gray shapes (target) according to the top-left reference pattern (source) shows a continuous, step-by-step spatial diffusion of color features from source to target. 
  \textbf{(D) Backtracking in Sudoku-Extreme:} on the hardest Sudoku boards (\textit{tdoku} difficulty 10; Figure S\ref{figure_sudoku_full_tts}B), cells demonstrate collective trial-and-error: they first reach a near-valid grid but with a few conflicting digits ($t = 60$), they escape that local minimum by temporarily increasing constraint violations ($t = 116$), and finally converging to a correct global consensus ($t = 250$). 
  }
  \label{figure_reasoning_skills}
\end{figure*}

\subsection{NCAs can solve complex reasoning tasks with a compact, fully local architecture}
\label{subsec_compact_local_reasoners}

On all the evaluated visual reasoning benchmarks (Mazes, Sudoku, ARC-AGI-1), NCAs demonstrate strong accuracy using a highly compact, local architecture with few parameters (Table~\ref{table_results_across_benchmarks}). Parameter efficiency is achieved by weight sharing: identical, local rules are executed recurrently over space and time to produce complex, multi-step reasoning. Because of the locality constraints, information can only spread between neighboring cells, which introduces spatial propagation delays. As such, NCAs necessarily require more iterations (higher $D$) to reach the correct solution than their globally connected counterparts. Yet, we find that they maintain a modest total FLOP budget (Table~\ref{table_results_across_benchmarks}) while supporting parallel, asynchronous execution at inference. 

In tasks where local information is key to reasoning, such as mazes, we find that NCAs engage in a form of distributed spatio-temporal reasoning over different potential paths (Figure ~\ref{figure_reasoning_skills}A,B), qualitatively similar to exploration by slime molds~\citep{Nakagaki2000} or breadth-first search~\citep{Earle2023}. 
Interestingly, recurrent models without locality constraints, such as TRM, do not appear to follow a localized path (Figure S\ref{figure_trm_solving}) and typically converge in fewer steps using non-local jumps (Table S\ref{table_locality_bottleneck}). NCAs instead generate a form of interpretable \enquote{spatial chain-of-thought} through visible, step-by-step traces of spatial reasoning in mazes.
This local information flow in reasoning can also be observed in the ARC-AGI-1 tasks, where NCAs appear to propagate image information gradually across the grid to solve the task (Figure~\ref{figure_reasoning_skills}C).

Even for tasks like Sudoku, where global constraints are critical, we find that the locality bottleneck does not prevent NCAs from discovering valid solutions. In fact, even on Sudoku-Extreme boards requiring extensive search and backtracks, NCAs are able to escape states where local constraints are satisfied but global ones are not, and to iterate until arriving at a solution that eventually respects the global constraints (Figure~\ref{figure_reasoning_skills}D). 

\begin{figure}[h!]
  \centering
  \includegraphics[width=0.48\textwidth]{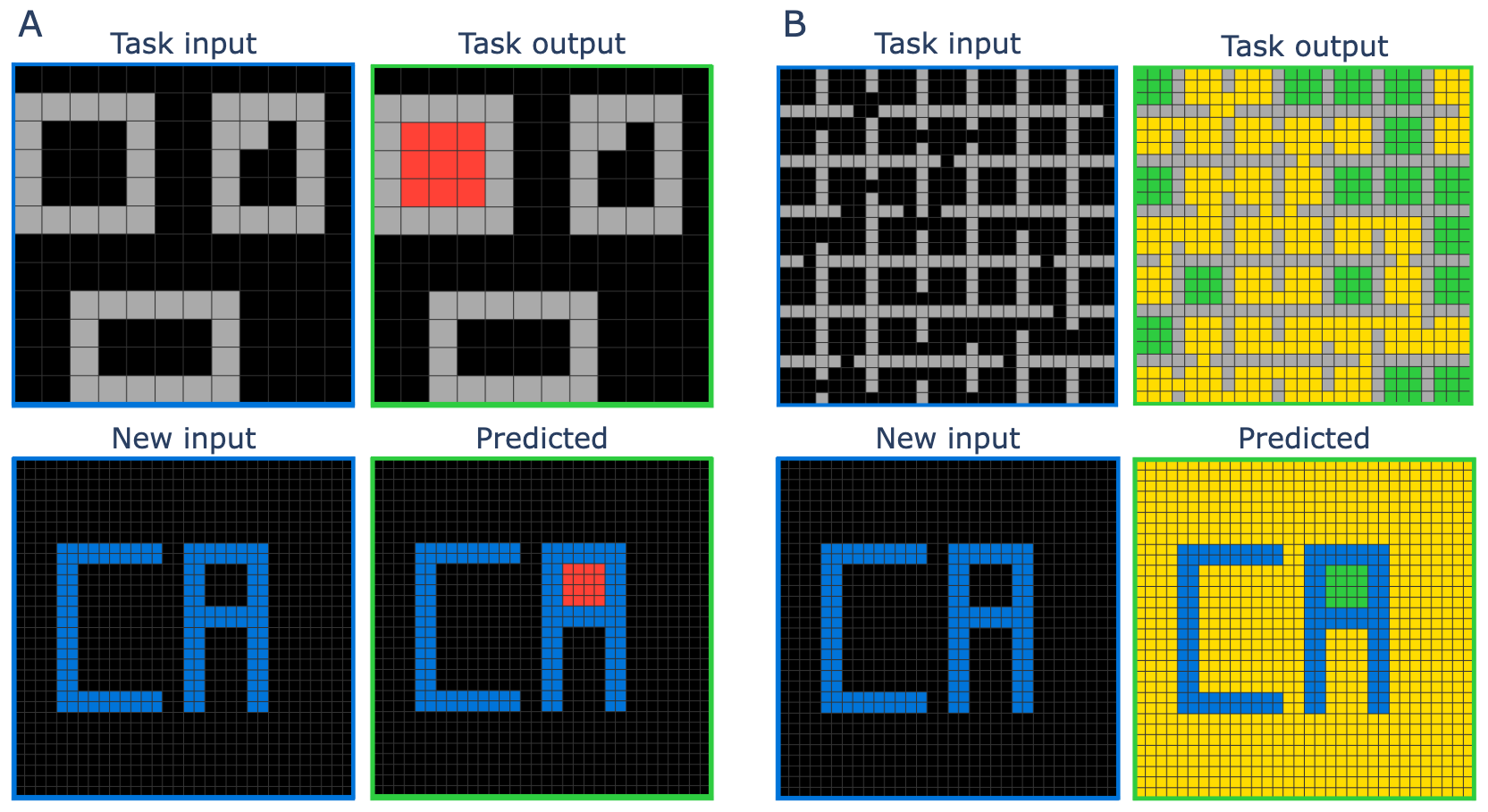}
\caption{\textbf{Multi-task execution in a single NCA.} When evaluating the pre-trained NCA on an unseen input (\enquote{CA} in blue), conditioned under different task embeddings, the model applies the specific transformation corresponding to each task (filling enclosed and/or open regions with target color), demonstrating context-dependent execution rather than input memorization.}
  \label{fig_arc_task}
\end{figure}

We find that the locality constraints of NCAs do not limit them to a single mode of spatio-temporal reasoning dynamics, either. A single NCA model can transfer across ARC-AGI-1 tasks, engaging in distinct forms of spatio-temporal reasoning when conditioned on unique, learnable task embeddings. When tested on novel, unseen inputs, the task embedding acts as a high-level program instruction, such that the NCA executes the relevant spatial rule until achieving the desired input-output transformation (Figure S\ref{fig_arc_task}).


\subsection{Scaling compute in NCAs enables generalization to harder tasks} 
\label{subsec_ood_generalization}

We next examined whether NCAs can generalize out-of-distribution to harder versions of the tasks they were trained on if we scale up compute at test-time. We used multiple ways to scale compute: launching independent, stochastic rollouts with varying initialization (parallel trials scaling), extending rollout length (temporal scaling), and expanding grid size (spatial substrate scaling).

\begin{figure*}
  \centering
  \includegraphics[width=\textwidth]{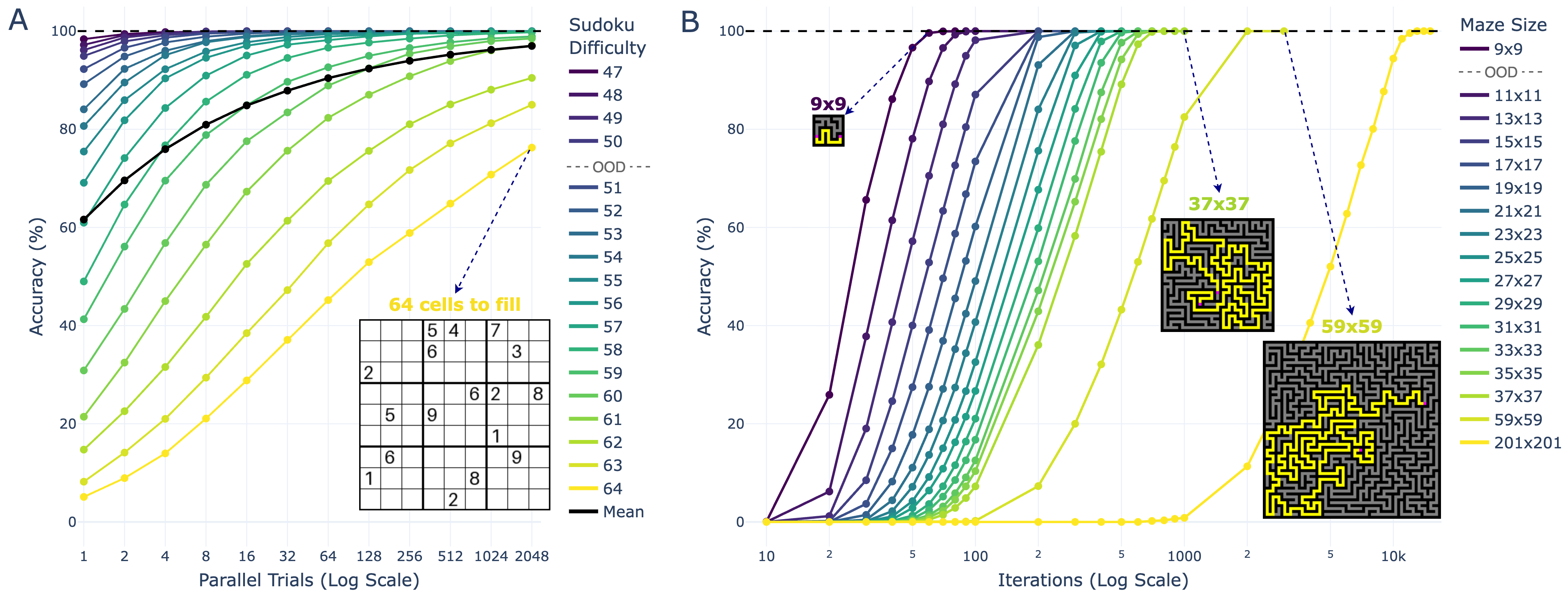}
  \caption{Test-time scaling improves NCA generalization on out-of-distribution (OOD) boards.
  \textbf{(A) Parallel state-space exploration:} running parallel, stochastic rollouts significantly improves OOD generalization on Sudoku.
  \textbf{(B) Spatial substrate scaling:} NCAs trained on $9\times9$ mazes generalize to up to $500\times$ larger mazes given additional substrate and iterations.
  Mean curves over 3 test seeds.
  }
  \label{figure_ood_generalization}
\end{figure*}

\begin{figure}[h!]
  \centering
  \includegraphics[width=0.5\textwidth]{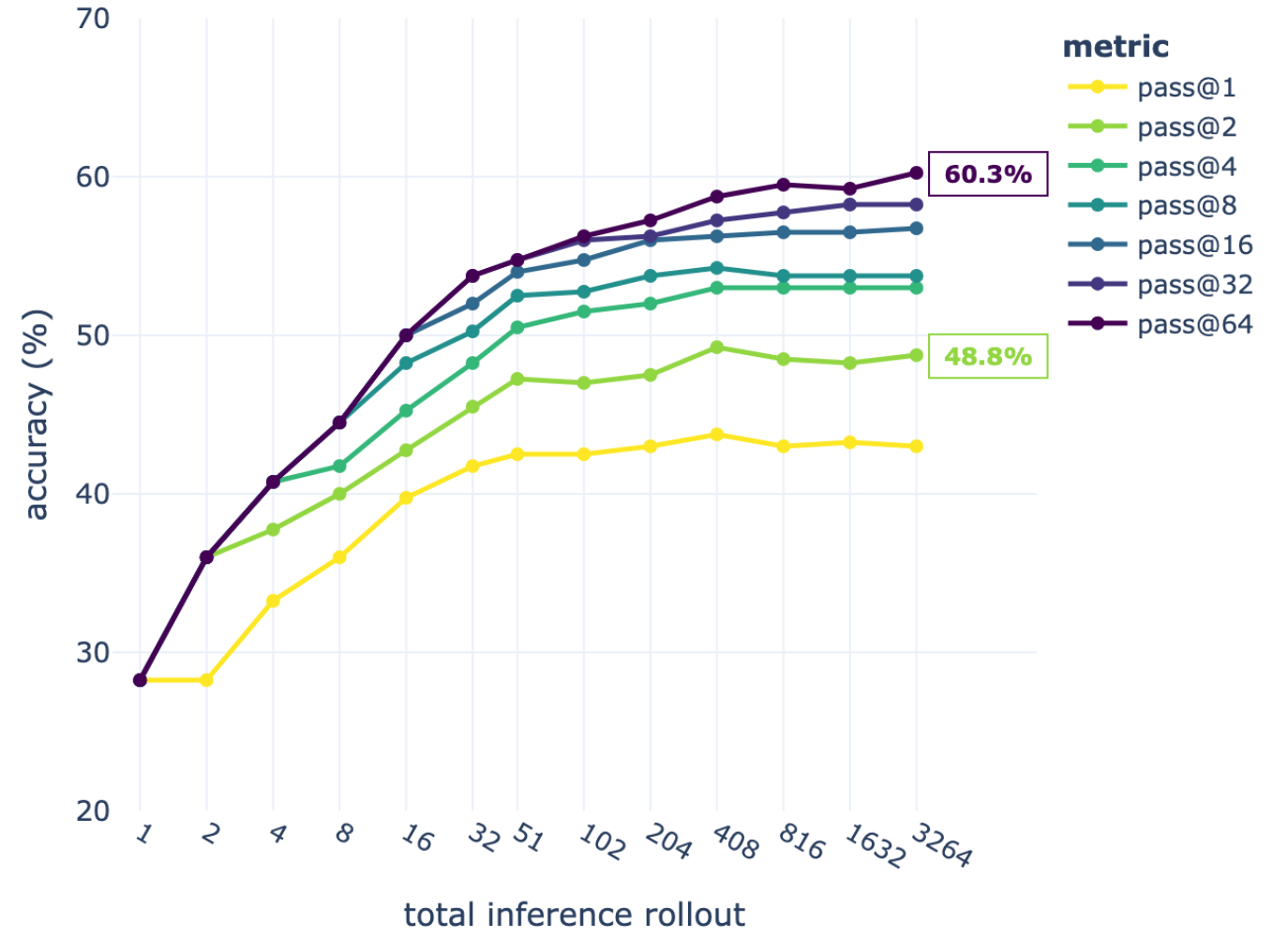}
    \caption{\textbf{Test-time compute scaling on ARC-AGI public evaluation set.} Pass@$K$ scaling under the offline-first-augmentation policy (see Appendix~\ref{subsec_training_arc_details} and~\ref{subsec_testing_details}), reaching 60.3\% Pass@64\protect\footnotemark.}
  \label{figure_arc_ttt}
\end{figure}

First, similar to other works in the literature, we find that running parallel, longer trials and selecting candidates via scoring significantly improves performance and OOD generalization. For instance, in Sudoku-OOD,  it unlocks solutions to Sudokus with as few as 17 clues despite being trained only on boards with at least 31 clues (Figure~\ref{figure_ood_generalization}A). Here test-time scaling acts as parallel state-space exploration: independent rollouts navigate distinct regions of the state space until convergence, generating several candidate solutions which the scoring mechanism can then select from. This simple exploration strategy shows similar improvements in performance for solving hard instances of Sudoku-Extreme (Figure S\ref{figure_sudoku_full_tts}B), Maze-Hard (Figure S\ref{figure_maze_full_tts}), and ARC-AGI-1 (Figure~\ref{figure_arc_ttt}).

\footnotetext{Using an NCA ensemble (NCA E) of 3 models independently fine-tuned further boosts performances to 63.0\%.}

\begin{figure}[h!]
  \centering
  \includegraphics[width=0.48\textwidth]{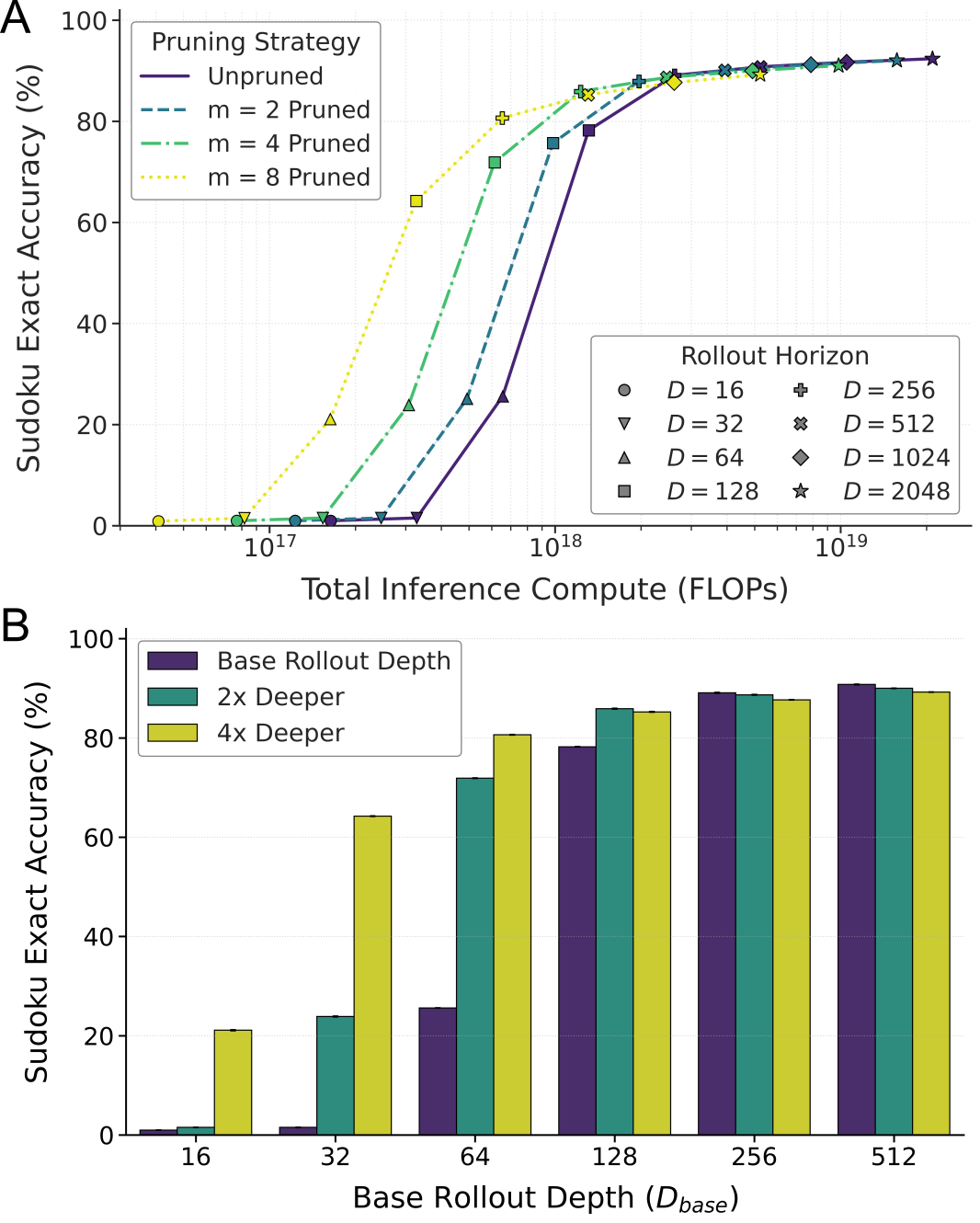}
  \caption{Scaling laws and iso-compute allocation of test-time pruning on Sudoku-Extreme. (A) Exact accuracy versus total FLOPs across rollout depths $D$ for unpruned baselines and pruning divisors $m$ (shaded bands indicate $\pm 1$ SD). (B) Allocation of fixed compute budgets to breadth (unpruned $D_\text{base}$) versus depth ($2\cdot D_{\text{base}}$ with $m=4$; $4\cdot D_{\text{base}}$ with $m=8$). Reallocating compute into depth yields accuracy gains up to $D_{\text{base}}=128$.}
  \label{figure_efficient_tts}
\end{figure}

Running parallel stochastic rollouts increases compute costs, of course, proportionally to $K \times D$. Under limited compute budgets, however, test-time scaling can be made much more efficient by pruning redundant rollouts. Using \textit{Niche-Capped Diversity Pruning} on Sudoku-Extreme, we find that the empirical compute-accuracy Pareto frontiers significantly shift to the left as the number of halving steps ($m$) increases: across the low-to-mid compute regimes, pruned rollouts achieve similar accuracy with up to $4\times$ lower compute compared to unpruned baselines (Figure~\ref{figure_efficient_tts}A). Similarly, at equivalent FLOP budgets, deeper rollouts compressed via pruning outperform standard baselines in compute-constrained regimes: at a base budget $D_{\text{base}} = 64$, quadrupling depth ($D=256, m=8$) boosts accuracy from $25.6\%$ to $81.7\%$ (Figure~\ref{figure_efficient_tts}B). Past a certain budget ($D_{\text{base}} = 256$), pruning yields only small differences over full parallel sampling.

On Maze-OOD, we also investigate length generalization via a second form of test-time scaling: spatial substrate scaling. 
With neither weight updates nor architectural modifications, NCAs generalize on out-of-distribution mazes up to $500\times$ larger (Figure~\ref{figure_ood_generalization}B). Interestingly, the NCAs learned a seemingly exact, generalizable, and interpretable pathfinding algorithm with wavefront expansion and backtracking that scales efficiently across the expanded substrates. This leads to impressive length generalization despite being trained solely on small $9\times9$ mazes (3 minutes training on TPU).

\begin{figure*}
    \centering
  \includegraphics[width=\textwidth]{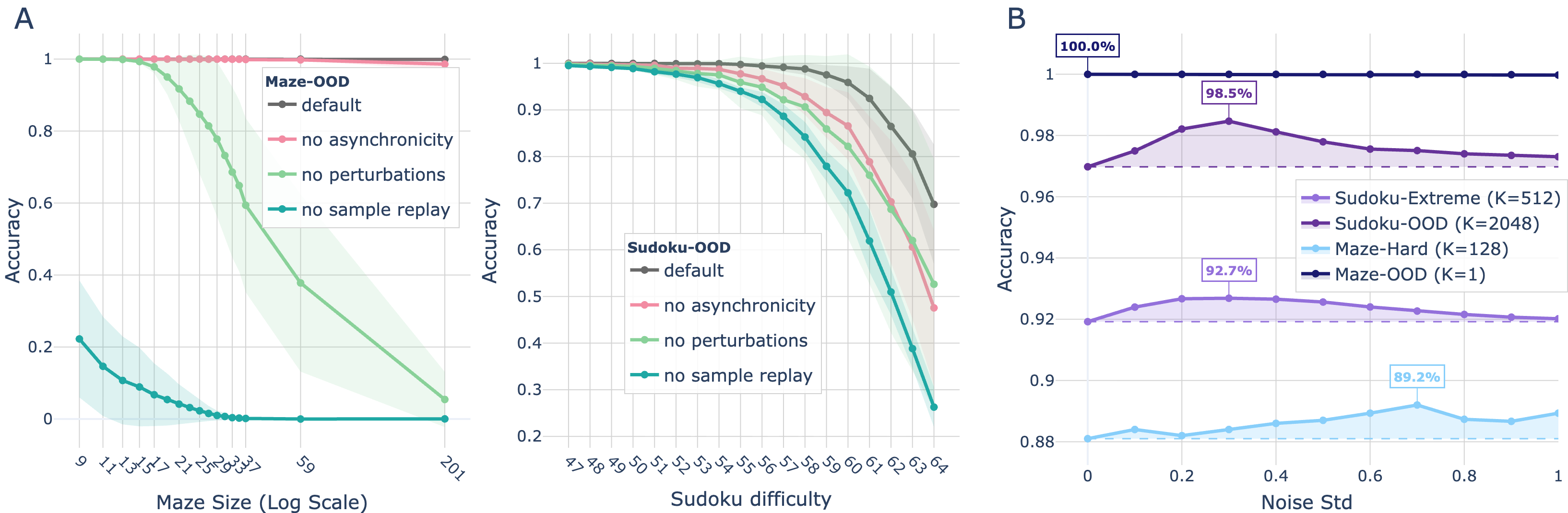}
  \caption{Stochasticity and perturbations are beneficial both at training and inference for generalization. \textbf{(A) Training ingredients ablation study:} the use of asynchronous updates (in particular for Sudoku), stochastic perturbations (noise, damage and target swap) as well as sample replay during training are critical ingredients for generalization to out-of-distribution instances. 
  Mean-std curves over 3 train seeds are displayed. \textbf{(B) Noise-injection at test time:} when adding noise of varying magnitude at test time, not only do the learned NCA rules show perfect robustness but noise even boosts performances across all benchmarks and noise scales. Metrics are averaged over 3 test seeds. }
  \label{figure_noise_generalization}
\end{figure*}

To better understand how NCAs can generalize to harder tasks we ablated various components of the training and test-time pipeline. We find that training perturbations combined with sample replay act as critical regularizers during training to achieve generalization (Figure \ref{figure_noise_generalization}A). By exposing the model to perturbed states outside standard solution paths and varying rollout depths $D$, these strategies generate a rich distribution of dynamic trajectories during training and foster the learning of general solutions. Interestingly, asynchronous updates are also critical for generalization in Sudoku, likely because updating cells non-simultaneously helps to break symmetries and escape local minima during constraint resolution. We also find that injecting random noise to cell states at test-time is not only tolerated (i.e. does not degrade performance) but even improves reasoning accuracy across benchmarks (Figure \ref{figure_noise_generalization}B), suggesting noise is a beneficial feature for NCAs, not a vulnerability.

\subsection{NCAs are robust adaptive reasoners}
\label{subsec_robust_adaptive_reasoners}

\begin{figure}
  \centering
  \includegraphics[width=0.48\textwidth]{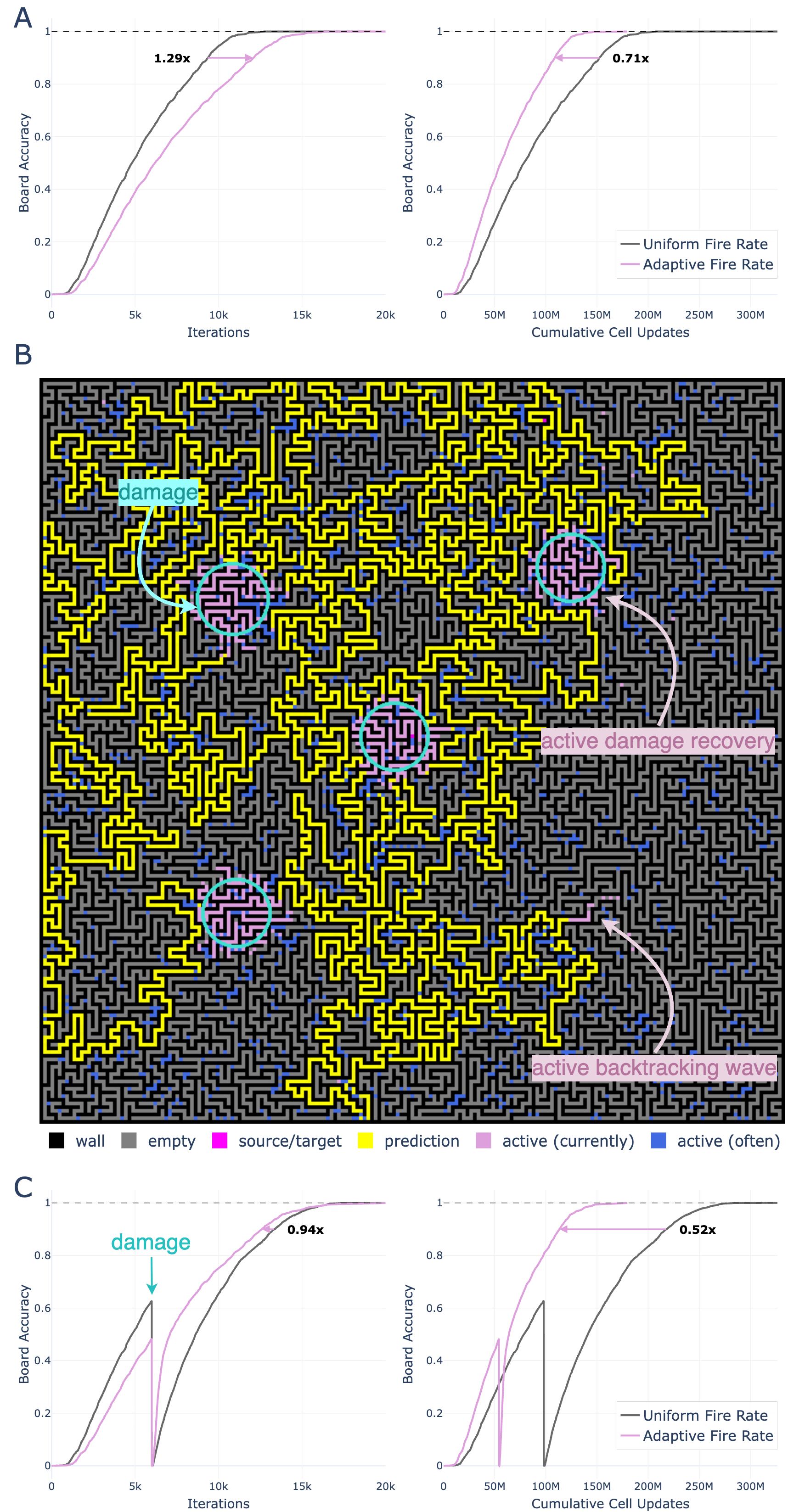}
  \caption{Robust adaptive compute on extra-large mazes. 
  \textbf{(A) Compute savings:} lowering the fire rate of confident cells ($p_{\text{fire}}$=0.4 if $c_{i,j}$>0.95, else 0.8) requires more steps to solve the mazes (reach y=0.95 with $1.29\times$ steps), but reduces total cell updates to $0.71\times$. 
  \textbf{(B) Localized activity:} 10 steps after adding damage mid-rollout (cyan circles), cells automatically activate either on damaged zones or on the yet-unsolved backtracking front (pink), while stable path segments remain largely dormant. 
  \textbf{(C) Enhanced damage recovery:} When injecting damage at $t=6000$, the adaptive strategy recovers faster ($0.94\times$ steps) and cuts cumulative cell operations nearly in half ($0.52\times$). Metrics are averaged over 3 test seeds.}
  \label{figure_maze_xl}
\end{figure}

One defining characteristic of NCAs is asynchronous execution: cells update independently without a shared global clock. While the experiments above apply a uniform update rate across the grid, the decentralized design of NCAs allows cells to decide locally when and how often to update.
We explored the use of such adaptive compute within NCA models trained on the Maze-OOD benchmark when deployed on extra large $201\times201$ grids. Specifically, rather than updating uniformly, each cell's update probability correlates with local certainty: the probability to update is low if the cell is \enquote{confident} ($p_{\text{fire}}=0.4$ if $c_{i,j}>0.95$), while uncertain cells continue to update at the default rate ($p_{\text{fire}}=0.8$).
We found that with this form of adaptive compute the NCAs required more global iterations to converge to the correct solutions, but, because each iteration involved fewer cell updates, the total number of cell updates required to converge was reduced by about 30\% (Figure \ref{figure_maze_xl}A). This simple strategy shows that compute can be effectively concentrated on active, unsolved regions of the problem space while saving resources elsewhere.

We then examined how adaptive compute impacted the robustness of NCAs in response to damage. Mid-rollout ($t=6000$), we damaged the NCAs by zeroing out cell states in random circular patches of the maze (Figure~\ref{figure_maze_xl}B, cyan circles). In response, adaptive updates naturally focused on either the damaged zones or on the end of the yet-unsolved backtracking front (pink segments). Stable path segments however remained largely dormant, saving compute resources when not needed.
Interestingly, this adaptive compute accelerated damage repair: whether measured in iterations or total cell updates, adaptive NCAs repaired damaged paths and converged to the correct solution more efficiently, cutting the total repair-and-solve time in half ($0.52\times$; Figure \ref{figure_maze_xl}C). These results suggest that fault tolerance (recovering from damage at a scale never seen during training) and efficiency go hand in hand, pointing to adaptive compute strategies as another potential path towards more efficient, large-scale reasoning.

\subsection{NCAs can scale to reason in pixel space}
\label{subsec_reasoning_in_pixel_space}

\begin{figure}
  \centering
  \includegraphics[width=0.48\textwidth]{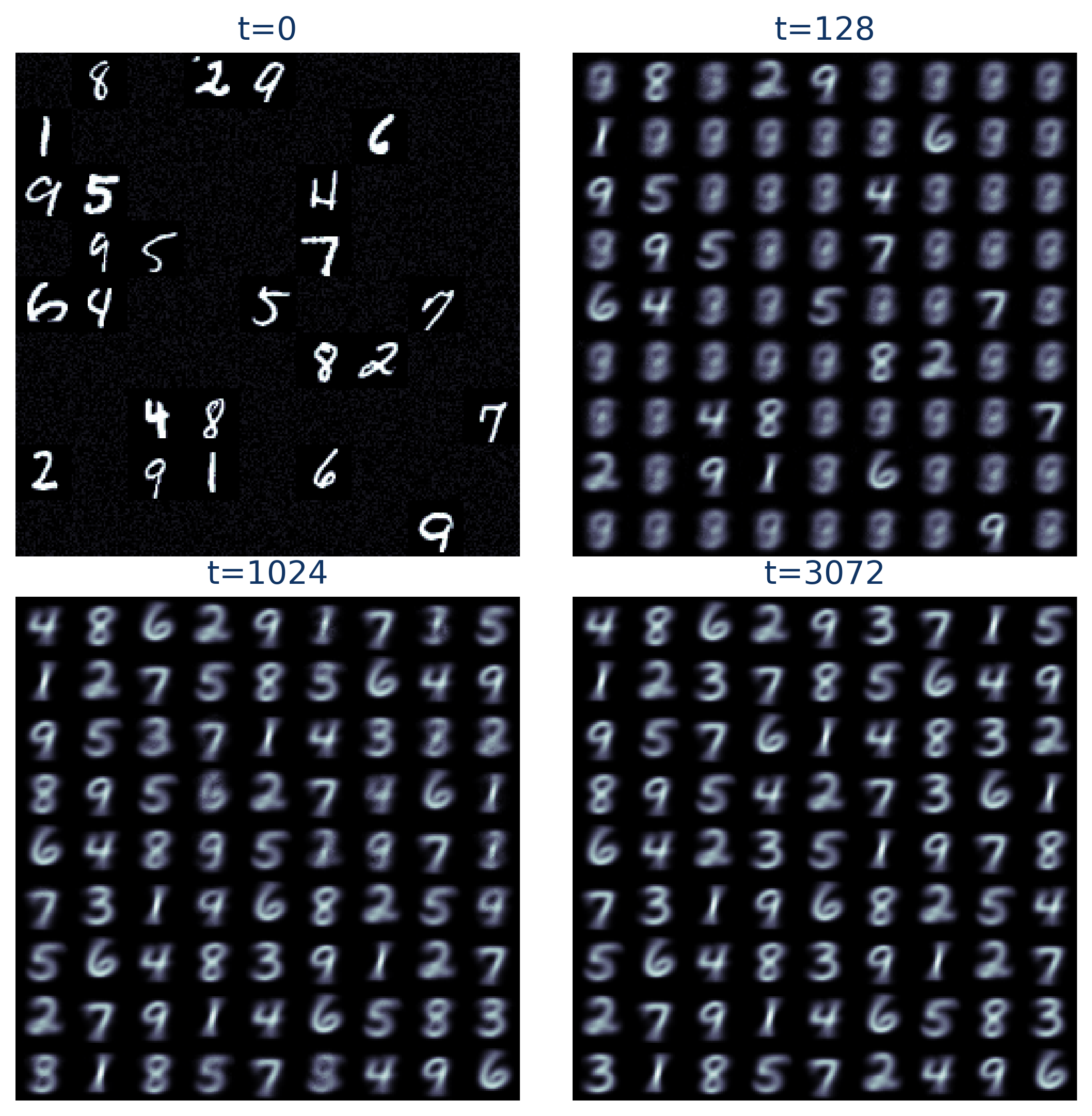}
  \caption{$256 \times 256$ NCA solving an out-of-distribution \textit{hard} sample from Visual Sudoku.}
  \label{visual_sudoku_arxiv}
\end{figure}

Our results thus far have demonstrated that NCAs can solve tasks where task-specific semantic information is encoded into the inputs and outputs of the model, and where training occurs on limited grid sizes (under $32 \times 32$ pixels). We now demonstrate that we can relax these constraints, removing task-specific semantic information, making the models perform iterative reasoning directly in pixel-space over very large grids. Specifically, we train NCAs on Visual Sudoku, a pure \enquote{image-to-image} reasoning task: presented with a $256 \times 256$ image of an incomplete Sudoku board  with randomly sampled MNIST digits as input, NCAs must return a completed, rendered image of the solved Sudoku. This task differs significantly from the previous ones, presenting a more general challenge as it requires (i) learning effective communication and coordination across hundreds of cells at training time; and (ii) simultaneously solving three separate tasks (image classification, Sudoku solving, and image rendering), all three at scales far beyond what an individual cell can perceive locally.

To that end, we scale an \enquote{off-the-shelf} NCA architecture, with pixel inputs/outputs and fixed Sobel perception filters, trained with MSE loss against random MNIST images whose categories match the reference solution. We relax task constraints and provide input clues via the initial image only (t=0), leaving these pixels mutable by the NCA. The model is trained on \textit{easy} puzzles from the Visual-Sudoku dataset, and tested on both \textit{easy} and \textit{hard}. On novel, unseen Sudoku images, it achieves an 87.8\% solve rate on \textit{easy} and up to 18.9\% on \textit{hard}, using test-time scaling via longer-rollouts than those seen at training (see Figure S\ref{figure_visual_tts}).
Interestingly, qualitatively investigating the model's behavior shows some form of distributed \enquote{chain-of-thought} process through time over the pixel space (Figure \ref{visual_sudoku_arxiv}). At first, cells appear to perform classification of input digits by converting each clue (given as new, unseen MNIST images) into a canonical form (resembling the mean of its digit class), while filling blank slots with what looks like the average of all MNIST digits ($t=128$). Then, cells proceed to iteratively solve the Sudoku, with partial guesses visible in the intermediate steps as rendered superpositions of possible digits ($t=1024$), until reaching a consensus on the final correct guesses likewise rendered in canonical form ($t=3072$). 

While our training pipeline on Visual-Sudoku is memory-intensive and leaves room for optimization (Appendix~\ref{subsec_visual_sudoku_settings}), these results provide a proof-of-concept that fully local, compact models can scale to much harder tasks while executing very efficiently at inference in a fully decentralized fashion.

\section{Discussion}
Our main contributions in this paper were fourfold. 
First, we have shown that spatial locality is not a barrier for complex multi-step visual reasoning: evaluated on a suite of challenging benchmarks including mazes, Sudoku, and ARC-AGI-1, compact NCAs successfully generate structured reasoning dynamics through local, asynchronous updates (\ref{subsec_compact_local_reasoners}). 
Second, we have shown that our core training recipe for NCAs (using sample replay and state perturbations), combined with the three axes of compute expansion for NCAs (spatial substrate, temporal dimension, and stochastic trials), enable strong out-of-distribution generalization on hard tasks (\ref{subsec_ood_generalization}). 
Third, we highlighted compelling properties of NCAs, demonstrating that self-repair can be coupled with dynamic compute allocation to concentrate computational resources along active reasoning fronts (\ref{subsec_robust_adaptive_reasoners}).
Finally, our results on Visual Sudoku suggest that this decentralized framework can scale to larger-scale visual reasoning directly in raw pixel space, requiring cells to communicate across long distances (\ref{subsec_reasoning_in_pixel_space}).

Taken together, these results suggest that enforcing stronger locality constraints from the ground up could offer a principled alternative for designing AI systems that naturally map to decentralized, self-organizing hardware and its critical requirements, such as low-power communication and fault tolerance.
There are, however, several limitations to the present work that could be addressed in future works.
First, while inference is completely decentralized, training remains non-local as we use truncated backpropagation through time, requiring global loss aggregation and error propagation. Future work could explore training over shorter truncated horizons, for instance by decomposing tasks into local denoising objectives via diffusion-based curricula~\citep{Drozdova2026}, or by exploring local learning rules that eliminate global gradient backpropagation altogether~\citep{Ernoult2019-sy,Bellec2020}.
Second, while exploring the NCA state space at test-time via stochastic rollouts significantly increased the likelihood of converging to the correct solution, this exploration strategy remains a simple random search. Future work may investigate active exploration strategies, such as curiosity-driven diversity search, to uncover the diverse reachable attractors of these systems more efficiently than random search~\citep{Reinke2019,etcheverry:tel-04504878}.
Finally, while strict locality comes with various useful properties regarding perspectives for hardware co-design, it inevitably introduces information propagation delays and bottlenecks to resolve long-range dependencies. Future work could investigate NCA-like architectures that are predominantly local but augmented with sparse, low-bandwidth, longer-range channels of communication, alongside evaluations of their physical trade-offs. 

While the exact trajectory of future computing hardware remains uncertain, emerging paradigms such as neuromorphic and distributed spatial processors highlight a need for AI architectures that can operate in a low-power, fault-tolerant, and inherently decentralized manner. We hope the insights from this study provide a modest step toward bridging that algorithmic and physical divide.

\bibliography{main}


\appendix
\noindent 
\textbf{\Large Appendix}
\vspace{1em}

\noindent\textbf{\ref{sec_benchmarks} Benchmarks}\dotfill\textbf{\pageref{sec_benchmarks}} \\
\hspace*{1.5em}\ref{subsec_maze_benchmark} Maze\dotfill\pageref{subsec_maze_benchmark} \\
\hspace*{1.5em}\ref{subsec_sudoku_benchmark} Sudoku\dotfill\pageref{subsec_sudoku_benchmark} \\
\hspace*{1.5em}\ref{subsec_arc_benchmark} ARC-AGI-1\dotfill\pageref{subsec_arc_benchmark} \\
\hspace*{1.5em}\ref{subsec_visual_sudoku_benchmark} Visual-Sudoku\dotfill\pageref{subsec_visual_sudoku_benchmark} \\[0.5em]
\noindent\textbf{\ref{sec_method_details} Method Details}\dotfill\textbf{\pageref{sec_method_details}} \\
\hspace*{1.5em}\ref{subsec_architecture_details} Architecture\dotfill\pageref{subsec_architecture_details} \\
\hspace*{1.5em}\ref{subsec_training_details} Training with Sample Replay \dotfill\pageref{subsec_training_details} \\
\hspace*{1.5em}\ref{subsec_training_arc_details} Training with Time Encoding (ARC)\dotfill\pageref{subsec_training_arc_details} \\
\hspace*{1.5em}\ref{subsec_testing_details} Evaluation Protocol\dotfill\pageref{subsec_testing_details} \\
\hspace*{1.5em}\ref{subsec_flops_estimation} FLOPs Estimation\dotfill\pageref{subsec_flops_estimation} \\
\hspace*{1.5em}\ref{subsec_tts_pruning_details} TTS Pruning Method\dotfill\pageref{subsec_tts_pruning_details} \\[0.5em]
\noindent\textbf{\ref{sec_experimental_settings} Experimental Settings}\dotfill\textbf{\pageref{sec_experimental_settings}} \\
\hspace*{1.5em}\ref{subsec_maze_settings} Maze\dotfill\pageref{subsec_maze_settings} \\
\hspace*{1.5em}\ref{subsec_sudoku_settings} Sudoku\dotfill\pageref{subsec_sudoku_settings} \\
\hspace*{1.5em}\ref{subsec_arc_settings} ARC-AGI-1\dotfill\pageref{subsec_arc_settings} \\
\hspace*{1.5em}\ref{subsec_visual_sudoku_settings} Visual-Sudoku\dotfill\pageref{subsec_visual_sudoku_settings} \\[0.5em]
\noindent\textbf{\ref{sec_additional_results} Additional Results}\dotfill\textbf{\pageref{sec_additional_results}} \\
\hspace*{1.5em}\ref{subsec_fixed_attention} Fixed Attention Analysis\dotfill\pageref{subsec_fixed_attention} \\
\hspace*{1.5em}\ref{subsec_locality_bottleneck} Impact of the Locality Bottleneck\dotfill\pageref{subsec_locality_bottleneck} \\
\hspace*{1.5em}\ref{subsec_tts_full_results} Extended Test-Time Scaling Results\dotfill\pageref{subsec_tts_full_results} \\
\hspace*{1.5em}\ref{subsec_tts_pruning_extended} Extended TTS Pruning Results\dotfill\pageref{subsec_tts_pruning_extended} \\

\vspace{0.7em}

\section{Benchmarks}
\label{sec_benchmarks}

\subsection{Maze}
\label{subsec_maze_benchmark}

We first evaluate the NCA's \textit{path-finding} abilities on two maze benchmarks: the \textbf{Maze-OOD} benchmark to test generalization to out-of-distribution maze sizes, and the \textbf{Maze-Hard} benchmark to test on hard mazes with multiple solutions.

\paragraph{Maze-OOD.} Dataset from \cite{Bansal2022} generated using the \texttt{easy-to-hard} python package data, which tests out-of-distribution spatial generalization. The training dataset contains 50K examples of small $9 \times 9$ grids (each with a unique solution). The test set contains larger mazes of sizes ranging from $9 \times 9$ to $37 \times 37$ (in increments of 2, 10K each), as well as extreme sizes $59 \times 59$ (10K) and $201 \times 201$ (1K) with very long dead-ends and deceptive paths. The vocabulary consists of $V=4$ tokens: empty cell ($0$), solution path ($1$), wall ($2$), and endpoints ($3$). Inputs specify wall and endpoint locations, and the NCA must complete the rest by predicting either empty or solution path. Training uses 8-way dihedral data augmentation (rotations and reflections).

\begin{figure}[h!]
  \centering
  \includegraphics[width=\linewidth]{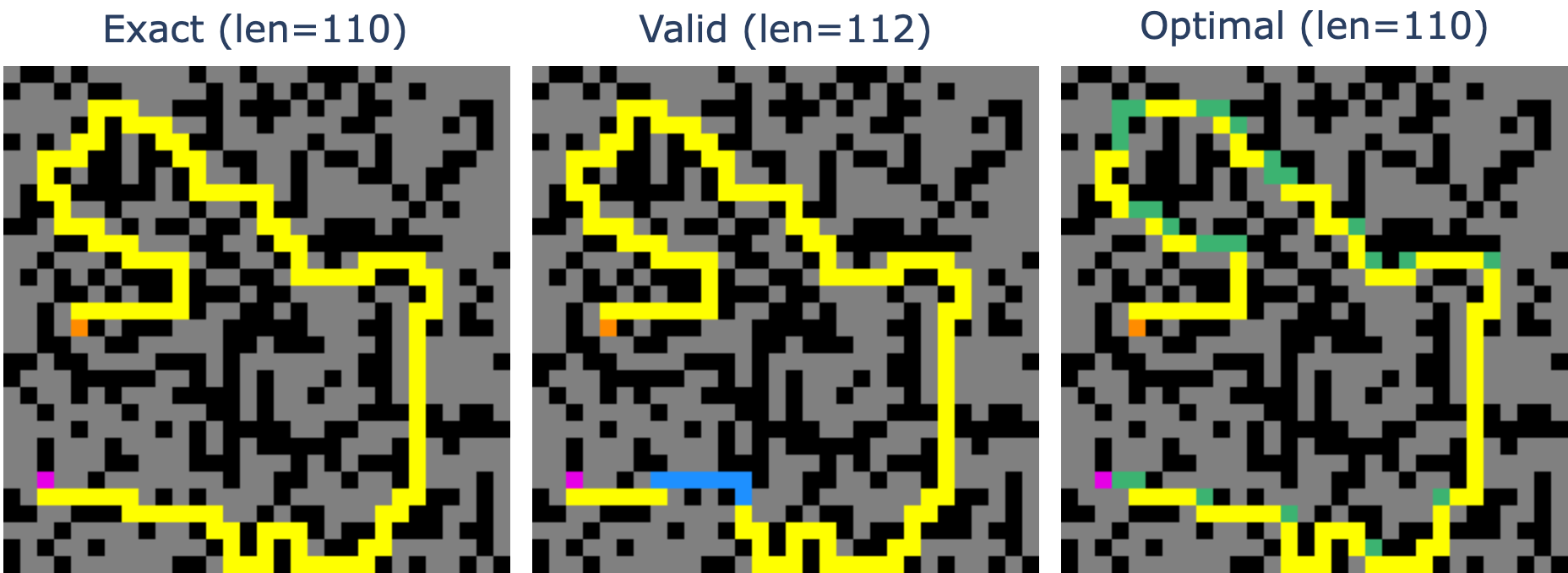}
  \caption{Maze-Hard task ambiguity. While trained on A* targets (\textit{exact}, left), there are multiple other \textit{valid}(middle, blue) and \textit{optimal} (right, green) solutions.}
  \label{figure_mhard_illustration}
\end{figure}

\paragraph{Maze-Hard.} Dataset from \cite{Wang2025} that tests the ability of a model to solve hard $30 \times 30$ mazes requiring solution paths of at least length 110. Unlike Maze-OOD, these mazes frequently contain multiple valid and optimal paths, introducing ambiguity against the single reference $\text{A}^*$ solution (Figure~\ref{figure_mhard_illustration}). The dataset contains 1000 training samples and 1000 test samples, with targets generated by the $A^*$ algorithm. The vocabulary consists of $V=5$ tokens: empty cell ($0$), solution path ($1$), wall ($2$), start ($3$), and goal ($4$) endpoints. Start and goal endpoints are explicitly distinguished this time because A* solutions (which this benchmark evaluates against) is asymmetric under start-goal swapping. Note that given this dataset models are trained to match $A^*$ solutions as opposed to learning \textit{any} optimal policy. We include this benchmark, in part, to test whether NCAs can learn $A^*$-like solutions via purely local dynamics, despite cells lacking the global distance-to-goal heuristic used in $A^*$. We also compare with training on the same dataset but with BFS-generated targets, with results also reported in Table~\ref{table_results_across_benchmarks}. Data augmentation is disabled for  $A^*$ (as it makes the supervised targets inconsistent) but used for BFS (with 8 dihedral transformations computed on the fly). We assess accuracy of the solutions based on whether the generated path is \textit{valid} (single continuous path which connects start and goal without crossing walls or branching), \textit{optimal} (valid path with minimal length) or \textit{exact} (matches the provided $A^*$/BFS solution). 

\subsection{Sudoku}
\label{subsec_sudoku_benchmark}

We then evaluate the ability of NCAs to solve a \textit{distributed constraint satisfaction problem} (DisCSP) on two Sudoku benchmarks: the \textbf{Sudoku-OOD} benchmark to test generalization to out-of-distribution Sudoku difficulties, and the \textbf{Sudoku-Extreme} benchmark to test on very hard Sudoku boards that require extensive \enquote{guesses} and \enquote{backtracks} to be solved.

\begin{figure}[h!]
  \centering
  \includegraphics[width=\linewidth]{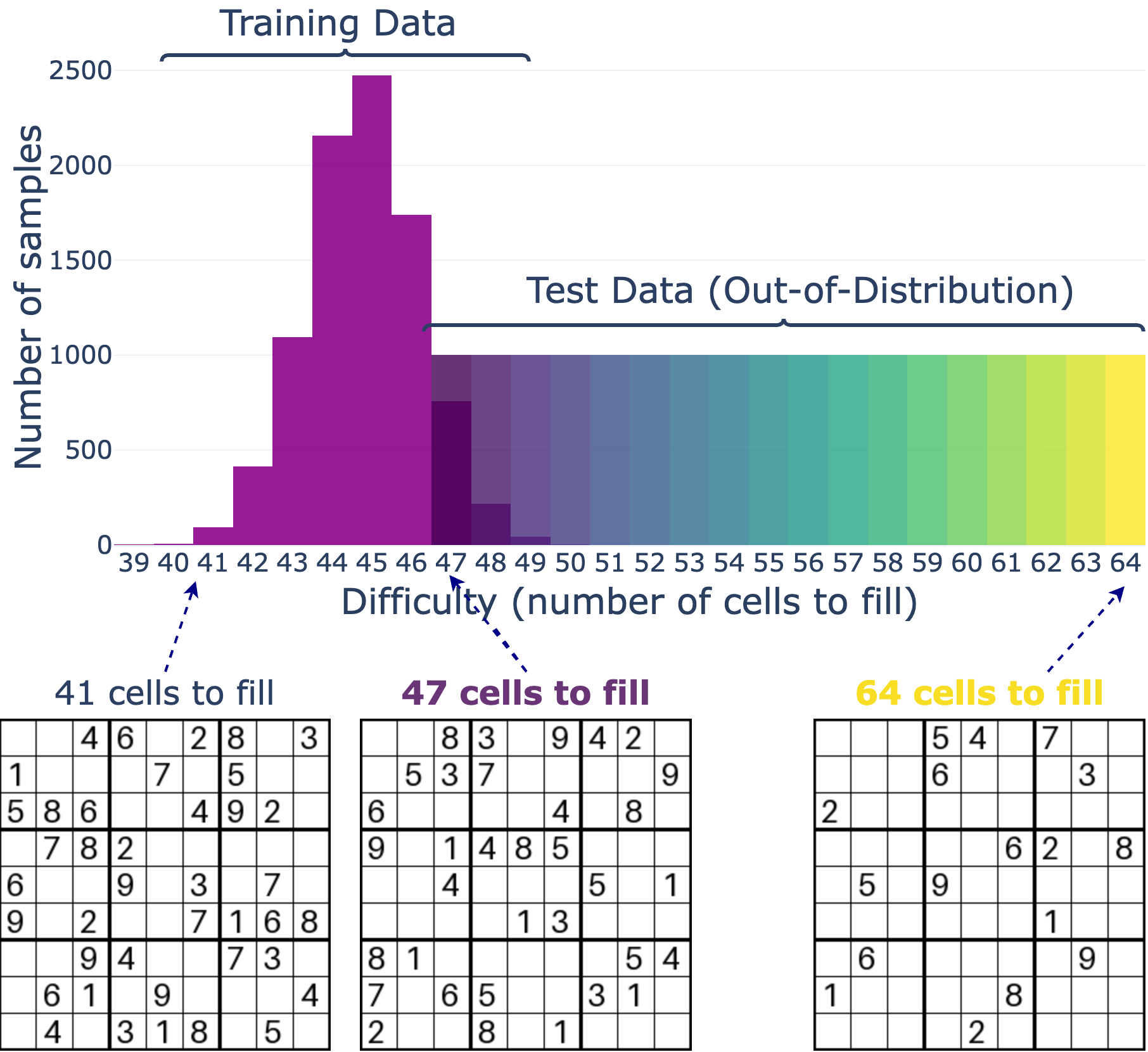}
  \caption{Sudoku-OOD dataset: the test set (47–64 cells to fill) presents a severe OOD shift relative to training (39–50 cells to fill)}
  \label{figure_sood_illustration}
\end{figure}

\paragraph{Sudoku-OOD.} Benchmark by \cite{Miyato2024} to test sudoku solving and out-of-distribution difficulty generalization. The training set contains 9K easy boards (and 1K validation boards) which were used in the SAT-Net paper \citep{Wang2019}. The test set contains harder boards across 18 difficulty levels (1K examples per level) which were used in the RRN paper \citep{Palm2017}, with a severe OOD shift relative to training (Figure~\ref{figure_sood_illustration}). Difficulty here is defined by the number of empty cells, ranging from 47 to 64 empty cells. The vocabulary consists of $V=10$ tokens: empty cell ($0$) and digits $1\text{--}9$. Input cells specify given prefilled digits, and the NCA must complete the board by predicting digits for all remaining empty cells. Training uses Sudoku-invariant symmetry data augmentation: transposition, random permutation of digits $1\text{--}9$, and random permutations of rows/columns within $3 \times 3$ blocks and of the blocks themselves.

\begin{figure}[h!]
  \centering
  \includegraphics[width=\linewidth]{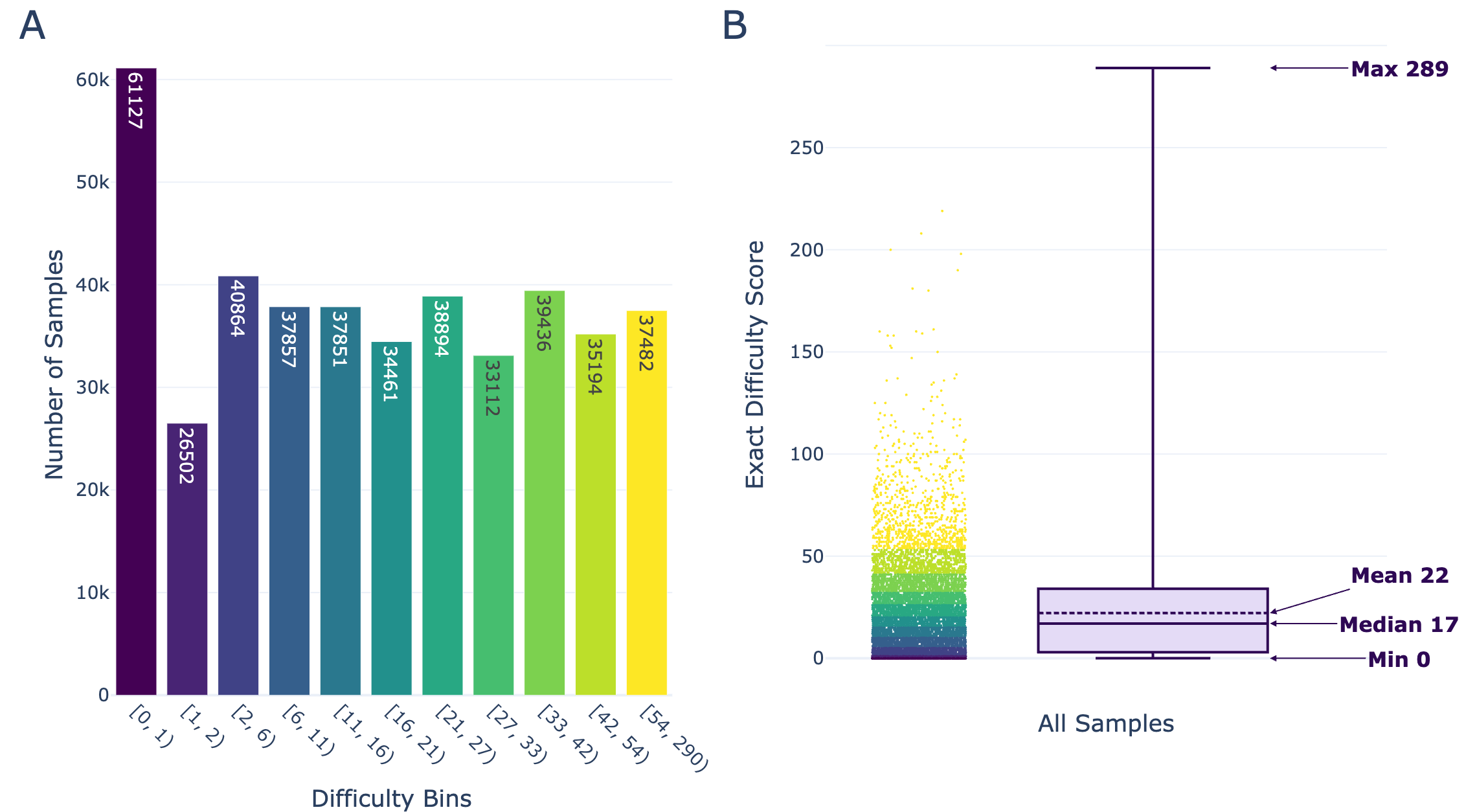}
  \caption{Sudoku Extreme Difficulty. (A) We group test samples in bins of increasing difficulties to plot results across difficulties in Figure~\ref{figure_sudoku_full_tts}. (B) Statistics of difficulties. }
  \label{figure_sext_illustration}
\end{figure}

\paragraph{Sudoku-Extreme.} Benchmark by~\cite{Wang2025} to test sudoku solving on extremely challenging boards requiring extensive search and backtracking. The training set contains only 1000 examples, and the test set contains 422,780 boards with difficulty scores ranging from 0 to 289, defined here as the number of backtracks needed by the logic-based \texttt{tdoku} solver\footnote{\href{https://t-dillon.github.io/tdoku/}{https://t-dillon.github.io/tdoku/}}(Figure~\ref{figure_sext_illustration}). Difficulties, i.e. numbers of backtracks needed to solve these boards, are significantly higher than the ones of other Sudoku datasets used in the literature~\citep{Wang2025}. As in Sudoku-OOD, the vocabulary consists of $V=10$ tokens ($0$ for empty cells, $1\text{--}9$ for digits) and training uses the same data augmentations.

\subsection{ARC-AGI-1}
\label{subsec_arc_benchmark}

In order to investigate whether conditioning enables a single NCA model to generalize across multiple tasks, we evaluate NCAs on the \textit{Abstraction and Reasoning Corpus} (ARC-AGI-1,~\cite{Chollet2019}), a benchmark designed for few-shot visual reasoning. Each task consists of a small set of input-output grid demonstrations (typically 2–5 examples) requiring the model to infer an underlying transformation rule and apply it to a novel test input. Following the training pipeline of recent works~\citep{Hu2025, Shu2026}, we complement the original training set with synthetic tasks generated by RE-ARC, which expands the set of available input-output pairs for the training tasks while preserving the core visual logic.

\subsection{Visual Sudoku}
\label{subsec_visual_sudoku_benchmark}

In Visual Sudoku, we investigate whether NCAs can be scaled to more complex problems on a much larger grids and perform \textit{reasoning in raw pixel space}. Visual-Sudoku is purely an image-to-image task, requiring to parse raw image pixels, classify the given digits, solve the DisCSP implementing the Sudoku, and then render an image representation of the correct digit as output. More specifically, Visual-Sudoku is a dataset derived from the \textit{easy} subset of Sudoku-OOD, rendering each puzzle pair as a pair of \textit{images} where each digit is represented as an independently randomly sampled $28 \times 28$ MNIST image from the corresponding class. In other words, a $1$ Sudoku clue is represented by randomly sampling an MNIST image of a $1$. Since each digit image is sampled independently, a $1$ in one part of the board appears differently from a $1$ in another, and likewise a $5$ given as an input clue will appear differently from the corresponding $5$ in the same position in the solved image. Thus, the pair of images will be one image for the unsolved state with clues, and one image for the solved state with all digits filled. This yields a $256 \times 256$ input image (adding 4 pixels of zero-padding to each side), with no additional semantic information provided. 

The training set uses the $9,000$ puzzles from the \textit{easy} set of Sudoku-OOD, and we test on both the in-distribution \textit{easy} test-set ($1000$ puzzles) and the \textit{hard} test-set ($1000$ puzzles sampled randomly). The digit images during training are sampled from the MNIST \textit{test} set, and the digit images during testing are sampled from the MNIST \textit{train} set (accidental inversion which we left as is, since both subsets remain strictly disjoint). To check Sudoku correctness at inference, we read out predictions using a non-learned classifier, which simply computes the pixel-wise $L_2$ distance between each $28 \times 28$ output patch and the mean image of each MNIST digit class, and predicts the closest class.

\section{Method}
\label{sec_method_details}
This section describes the model architecture, training procedure and test-time scaling evaluation protocol.

\subsection{Architecture}
\label{subsec_architecture_details}

The NCA operates on a 2D grid of size $H \times W$. Each grid cell at spatial coordinates $(i, j)$ contains a continuous state vector $x_{i,j} \in \mathbb{R}^C$, where $C$ is the total number of channels. The overall grid state at discrete time step $t$ is denoted by $X^{(t)} \in \mathbb{R}^{H \times W \times C}$.

\paragraph{Task Embedding.}
To project the discrete task representation (2D grid of tokens + task id for ARC) into the continuous latent space of the NCA we construct the following embeddings:
\begin{itemize}[leftmargin=12pt]
    \item[--] \textbf{Token Embedding:} Each discrete token $v \in \{0, 1, \dots, V-1\}$ (where $V$ is the vocabulary size) is represented by a $C_{\text{in}}$-dimensional vector $e_v \in \mathbb{R}^{C_{\text{in}}}$. These embeddings are fixed and predefined as orthonormal vectors for Maze, Sudoku, and ARC. For Visual Sudoku we directly use pixel values ($C_{\text{in}}=1$).
    \item[--] \textbf{Task Embedding (ARC):} a global task embedding $e_{\text{task}} \in \mathbb{R}^{C_{\text{task}}}$ is learned to condition the initialization state of NCA cells on the underlying task.
\end{itemize}

\paragraph{State Initialization.}
The initial continuous grid state $X^{(0)} \in \mathbb{R}^{H \times W \times C}$ is constructed as follows:
\begin{itemize}[leftmargin=12pt]
    \item[--] \textbf{Input Cells:} For cells where tokens are provided in the initial board (e.g., walls in Maze or given digits in Sudoku), the first $C_{\text{in}}$ channels are set to the corresponding token embedding $e_v$, while the remaining channels are initialized with Gaussian noise $\mathcal{N}(0, \sigma_{\text{init}}^2)$. During rollout, the $C_{\text{in}}$ channels remain strictly fixed to their initial token embeddings ($\Delta x_{i,j, \text{in}}^{(t)} = 0$), acting as immutable elements.
    \item[--] \textbf{Empty/Unsolved Cells:} For empty cells, all $C$ state channels are initialized with Gaussian noise $\mathcal{N}(0, \sigma_{\text{init}}^2)$.
    \item[--] \textbf{Task Conditioning (ARC):} For ARC, the global task embedding $e_{\text{task}}$ is injected into a dedicated channel slice of every cell across the grid.
    \item[--] \textbf{Input and Output padding (ARC):} To handle variable task dimensions, grids are padded into a $32 \times 32$ canvas using a new padding token which is explicitly masked out of the loss function.
    
\end{itemize}

\paragraph{Perception Module.}
At every step $t$, each cell $(i, j)$ inspects its local neighborhood $\mathcal{N}(i, j)$ (a $3 \times 3$ Moore neighborhood containing 9 cells) to extract a perception vector $z_{i,j}^{(t)}$:
\begin{itemize}[leftmargin=12pt]
    \item[--] \textbf{Convolutional Sensing (Maze, Visual Sudoku):} A set of $K_{\text{heads}}$ $3 \times 3$ convolution kernels $\{W_{\text{perc}}^{(k)}\}_{k=1}^{K_{\text{heads}}}$ is applied to the neighbor states. For each head $k$, the 9 neighbor states are linearly combined:
    \begin{equation*}
        z_{i,j}^{(k)} = \sum_{n=1}^{9} W_{\text{perc}, n}^{(k)} x_{n}^{(t)} \in \mathbb{R}^C
    \end{equation*}
    where $x_n^{(t)} \in \mathcal{N}(i, j)$. Concatenating all $K_{\text{heads}}$ outputs produces the perception vector $z_{i,j} \in \mathbb{R}^{K_{\text{heads}} \cdot C}$. For Maze, these kernels are learned. For Visual-Sudoku, they are kept fixed as the identity, vertical, and horizontal Sobel filters as in~\cite{Mordvintsev2020}.
    \item[--] \textbf{Fixed-Attention Sensing (Sudoku, ARC):} A set of $K_{\text{heads}}$ position-specific $3 \times 3$ attention matrices $\{A_{i,j}^{(k)}\}_{k=1}^{K_{\text{heads}}}$ and value projection matrices $\{V^{(k)}\}_{k=1}^{K_{\text{heads}}}$ (where $V^{(k)} \in \mathbb{R}^{\frac{C}{K_{\text{heads}}} \times C}$) is applied to the neighbor states. Each head $k$ projects and linearly combines neighbor states:
    \begin{equation*}
        z_{i,j}^{(k)} = \sum_{n=1}^{9} A_{i,j, n}^{(k)} \left(V^{(k)} x_{n}^{(t)}\right) \in \mathbb{R}^{\frac{C}{K_{\text{heads}}}}
    \end{equation*}
    where $x_n^{(t)} \in \mathcal{N}(i, j)$. Concatenating all $K_{\text{heads}}$ outputs produces the perception vector $z_{i,j} \in \mathbb{R}^C$.
\end{itemize}

\paragraph{Update Module.}
The update module is shared across all cells and maps the perception vector $z_{i,j}^{(t)}$ to a proposed state update $\Delta x_{i,j}^{(t)}$.
\begin{itemize}[leftmargin=12pt]
    \item[--] \textbf{MLP (Maze, Sudoku, Visual Sudoku):} A standard two-layer MLP with $\operatorname{ReLU}$ activation:
    \begin{equation*}
        \Delta x_{i,j}^{(t)} = W_2 \operatorname{ReLU}\left(W_1 z_{i,j}^{(t)} + b_1\right) + b_2
    \end{equation*}
    \item[--] \textbf{SwiGLU (ARC):} A gated activation layer that splits the projected representation into gate $g$ and value $u$:
    $\begin{aligned}
        [g, u] &= W_1 z_{i,j}^{(t)} + b_1, \\ 
        \Delta x_{i,j}^{(t)} &= W_2 \left(\operatorname{swish}(g) \odot u\right) + b_2
    \end{aligned}$
    where $\operatorname{swish}(g) = g \cdot \sigma(g)$, and $\odot$ denotes element-wise multiplication.
\end{itemize}
$W_1$ expands the perception vector by an expansion factor $E$, and $W_2$ projects back to $C$ channels.

\paragraph{Asynchronous Stochastic Execution.}
To simulate asynchronous cellular behavior, cells update stochastically according to a binary mask $m_{i,j}^{(t)} \in \{0, 1\}$. If $m_{i,j}^{(t)} = 0$, cell $(i, j)$ does not update at step $t$. We consider two firing strategies:
\begin{itemize}[leftmargin=12pt]
    \item[--] \textbf{Uniform Firing (Default):} Cells fire with a constant cell fire rate $p_{\text{fire}}$:
    \begin{equation*}
        m_{i,j}^{(t)} \sim \operatorname{Bernoulli}(p_{\text{fire}})
    \end{equation*}
    \item[--] \textbf{Adaptive Firing (Section~\ref{subsec_robust_adaptive_reasoners}):} The fire rate drops to $p_{\text{low}} < p_{\text{fire}}$ once a cell's confidence $c_{i,j}^{(t)}$ exceeds threshold $\tau_{\text{halt}}$:
    \begin{equation*}
        m_{i,j}^{(t)} \sim \operatorname{Bernoulli}\left(
        \begin{cases}
            p_{\text{low}} & \text{if } c_{i,j}^{(t)} > \tau_{\text{halt}}, \\
            p_{\text{fire}} & \text{otherwise}
        \end{cases}
        \right)
    \end{equation*}
\end{itemize}

\paragraph{Residual State Update.}
The cell state is updated via a residual connection:  $x_{i,j}^{(t+1)} = x_{i,j}^{(t)} + m_{i,j}^{(t)} \cdot \Delta x_{i,j}^{(t)}$. For Sudoku and ARC, we also use  
\textbf{per-cell group normalization} following \cite{Miyato2024}: cell states are seen as \textit{oscillators}, i.e. $U$ independent unit vectors of size $C / U$ that rotate on a sphere, and each unit is re-normalized after each update to unit $L_2$ norm. 

\paragraph{Prediction Head.}
At any step $t$, the predicted token $\hat{v}_{i,j}$ of cell $(i, j)$ is inferred from its $C_{\text{out}}$-dimensional readout slice $x_{i,j, \text{out}}$ by computing a score $\ell_{i,j}(v)$ for each candidate token $v$:
\begin{itemize}[leftmargin=12pt]
    \item[--] \textbf{$L_2$ Similarity (Maze):} Computes the inverse Euclidean distance between the cell readout slice and token embeddings:
    \begin{equation*}
        \ell_{i,j}(v) = \frac{1}{1 + \|x_{i,j, \text{out}} - e_v\|_2}
    \end{equation*}
    \item[--] \textbf{Cosine Similarity (Sudoku):} Computes the normalized dot product with token embeddings:
    \begin{equation*}
        \ell_{i,j}(v) = \frac{x_{i,j, \text{out}} \cdot e_v}{\|x_{i,j, \text{out}}\|_2 \|e_v\|_2}
    \end{equation*}
    \item[--] \textbf{MLP Head (ARC):} Projects the readout slice to class logits via a learned linear transformation:
    \begin{equation*}
        \ell_{i,j}(v) = \left(W_{\text{pred}} x_{i,j, \text{out}} + b_{\text{pred}}\right)_v
    \end{equation*}
     \item[--] \textbf{Parameter-free classifier (Visual Sudoku):} The output image patch for Sudoku cell $(i, j)$ is $I_{i,j} = X_{28i:28(i+1),\, 28j:28(j+1),\, \text{out}}$. 
     We compute predictions against empirical class-mean prototypes $P_v = \frac{1}{N_v} \sum_{k=1}^{N_v} I_v^{(k)}$, computed across the MNIST test set. Thus, score is:
    \begin{equation*}
        \ell_{i,j}(v) = - \|I_{i,j} - P_v\|_F^2
    \end{equation*}
\end{itemize}
The predicted token is $\hat{v}_{i,j} = \arg\max_v \ell_{i,j}(v)$.

\paragraph{Confidence Head.}
A per-cell scalar confidence $c_{i,j} \in [0, 1]$ can also be extracted at inference at any step $t$, using 
the maximum score across all candidate tokens:
    \begin{equation*}
        c_{i,j} = \max_v \ell_{i,j}(v)
    \end{equation*}
The global board confidence is the mean cell confidence across the grid:
\begin{equation*}
    c(X) = \frac{1}{H \cdot W} \sum_{i=1}^H \sum_{j=1}^W c_{i,j}
\end{equation*}

\paragraph{Loss Function.}
\begin{itemize}[leftmargin=12pt]
    \item[--] \textbf{Mean Squared Error (Maze, Sudoku, Visual Sudoku):} Measures the distance between the cell readout slice and target token embeddings:
    \begin{equation*}
        \mathcal{L}_{\text{MSE}} = \frac{1}{|\mathcal{W}| \cdot H \cdot W} \sum_{t \in \mathcal{W}} \sum_{i=1}^H \sum_{j=1}^W \left\|x_{i,j, \text{out}}^{(t)} - e_{y_{i,j}}\right\|_2^2
    \end{equation*}
    \item[--] \textbf{Cross-Entropy (ARC):} Standard categorical cross-entropy computed on the predicted class logits $\ell_{i,j}^{(t)}$ against target tokens:
    \begin{equation*}
        \mathcal{L}_{\text{CE}} = -\frac{1}{|\mathcal{W}| \cdot H \cdot W} \sum_{t \in \mathcal{W}} \sum_{i=1}^H \sum_{j=1}^W \log \operatorname{softmax}\left(\ell_{i,j}^{(t)}\right)_{y_{i,j}}
    \end{equation*}
\end{itemize}
The set of steps $\mathcal{W}$ that we compare with target within the chunk are either:
\begin{itemize}[leftmargin=12pt]
    \item[--] All-step supervision (Maze, Sudoku-OOD, Visual Sudoku): The loss is averaged across all $N_{\text{chunk}}$ steps of the sub-window ($|\mathcal{W}| = N_{\text{chunk}}$), providing stronger gradient signal but penalizing intermediate states along the trajectory.
    \item[--] Window-step supervision (ARC): The loss is evaluated across the final $w$ steps of the chunk ($1 < |\mathcal{W}| = w < N_{\text{chunk}}$), allowing early exploration while reinforcing stable convergence over the final trajectory.
    \item[--] Last-step Supervision (Sudoku-Extreme): The loss is evaluated only at the final step of the chunk ($|\mathcal{W}| = 1$), giving intermediate steps more freedom to explore solution space.
\end{itemize}

\subsection{Training with Sample Replay}
\label{subsec_training_details}

Following the original NCA training methodology~\citep{Mordvintsev2020}, we train our models using Backpropagation Through Time (BPTT) combined with a buffer of previously-inferred samples as well as training perturbations. The use of the replay buffer stabilizes long-horizon recurrent dynamics by emulating long execution trajectories without incurring the memory footprint of backpropagating through thousands of steps.

\paragraph{Training pipeline.}
A fixed-capacity buffer maintains $M$ state grids initialized with fresh board states (age 0). At each training iteration:
\begin{enumerate}
    \item[--] A batch of $B$ states is drawn uniformly at random from the sample replay buffer.
    \item[--] A fraction $r_{\text{seed}}$ of the sampled batch is replaced with fresh initial states (age 0).
    \item[--] Perturbations (target swapping, state noise, and damage) are applied to the batch.
    \item[--] The batch is unrolled for $N$ steps using the current NCA parameters.
    \item[--] A random sub-window of length $N_{\text{chunk}}$ is selected.
    \item[--] The loss is evaluated across the subwindow, gradients are computed with truncated BPTT on the chunk, and the NCA parameters are updated.
    \item[--] The final post-rollout states and their incremented ages are written back into the pool.
\end{enumerate}

\paragraph{Training Perturbations.}
Three types of perturbations are applied during training:
\begin{itemize}[leftmargin=12pt]
    \item[--] \textbf{State Noise (During Rollout):} At each rollout step, Gaussian noise $\mathcal{N}(0, \sigma_{\text{noise}}^2)$ is injected into cell states with temporal probability $p_{t}$ and spatial probability $p_{s}$ (excluding the frozen input channels of input cells).
    \item[--] \textbf{State Damage (Pre-Rollout):} With probability $p_{\text{damage}}$, $N_{\text{damage}}$ circular masks with radii $r \in [0.1, 0.4]$ (normalized to grid size) reset cell channels to zero.
    \item[--] \textbf{Target Swapping (Pre-Rollout):} With probability $p_{\text{swap}}$, a sample's input clues and target solution are replaced with those of another board from the dataset, forcing the NCA to constantly sense its environment and dynamically adapt its internal states toward the new task when such swap occurs.
\end{itemize}

\subsection{Training with time encoding (ARC)}
\label{subsec_training_arc_details}

Following prior work~\citep{Shu2026}, we condition model execution by injecting step-based temporal encodings. We hypothesize that solving ARC tasks requires multi-phase computation and that this temporal signal helps this staged reasoning. At step $t$, the step embedding $e_{\text{step}}^{(t)}$ is learned and broadcast across the grid and added directly to the hidden state slice of every cell prior to neighborhood sensing:
\begin{equation}
    x_{i,j, \text{hid}}^{(t)} \leftarrow x_{i,j, \text{hid}}^{(t)} + e_{\text{step}}^{(t)},
\end{equation}
where the time-encoding $e_{\text{step}}^{(t)}$ is a vector carrying no spatial coordinate or relational information.

At inference the global clock becomes an internal \emph{local clock}: each cell indexes the step embedding by the number of steps in which it has itself fired, keeping its state and counter frozen otherwise.

\paragraph{Canvas and online augmentation.}
 All grids are padded on a fixed $32\times32$ canvas~\citep{Shu2026}. At every optimization step the original input--target pair is rescaled by a random integer and placed at a random offset in the canvas, where a new border token marks the extent of the target grid. The rest of the canvas is filled with pad tokens that are masked out of the loss. This \textit{online} augmentation is identical in both the pre-training and the fine-tuning, so each pair is seen at a new scale and position every time it is sampled.

\paragraph{Pre-training.}
Following LoopViT~\citep{Shu2026}, we pre-train on training tasks augmented with the pairs from the RE-ARC dataset~\citep{Hodel2024}. At each training
step we proceed with the following pipeline: 1) sample a batch of input--target pairs; 2) render them on the canvas with fresh random scale and offset; 3) rollout $N$ steps and perform a training step. 

\paragraph{Test-time training (per task).}
For each evaluation task we build $51$ augmentations (\textit{offline} augmentations): the original plus five dihedral transforms, each with ten color permutations. The augmentation pipeline was built on previous work~\citep{Shu2026, Hu2025}. Every augmented variant receives its own task id, whose embedding is randomly initialized, while all remaining weights are loaded from the pre-trained model. We then fine-tune with the same loop as pre-training, sampling batches from the demonstration pairs pooled across the $51$ variants. The task embeddings and the model weights are updated jointly.

\subsection{Evaluation Protocol}
\label{subsec_testing_details}
At test time, inference consists of rolling out the trained NCA for $D$ iterations from an initial state $X^{(0)}$ initialized with Gaussian noise of scale $\sigma_{\text{init}}$. For test-time scaling experiments (section~\ref{subsec_ood_generalization}), we scale computation along two axes: increasing the rollout horizon $D$ (temporal scaling) and launching $K$ parallel rollouts with an automated selection mechanism (parallel trials scaling). Note that for Maze-OOD depth scaling is combined with scaling substrate size $S$ (spatial substrate scaling).

\paragraph{Depth Scaling ($D$).}
The model is unrolled for $D$ steps. Because information propagates locally at each step, scaling the iteration budget $D$ allows signals to traverse larger boards and perform iterative reasoning to solve more complicated tasks than the ones seen during training.

\paragraph{Width Scaling ($K$).}
To explore diverse solution trajectories for a given board, we run $K$ parallel rollouts starting from different initial states and running with stochastic asynchronous updates ($p_{\text{fire}} < 1$):
\begin{itemize}[leftmargin=12pt]
    \item[--] \textbf{Stochastic Initialization (Maze, Sudoku):} Each of the $K$ trials is initialized with independent Gaussian noise $\mathcal{N}(0, \sigma_{\text{init}}^2)$.
   
    \item[--] \textbf{Augmented Ensembles (ARC):} For each test image we aggregate
    predictions over a budget of $K = E \times A \times R$ rollouts:
    \begin{itemize}
        \item Offline augmentations ($A$ \emph{task augmented variants}): these are the same variants used for test-time training, so each test image is evaluated with its own fine-tuned task embedding;
        \item Online augmentations - $R$ stochastic rollouts per variant, which differ through the random hidden-state initialization, the asynchronous cell-firing mask, and a fresh random scaling and translation of the test input on the $32\times32$ canvas drawn independently for every rollout;
        \item Model ensemble - $E$ independently fine-tuned models generate $A \times R$ predictions which are aggregated in the same pool.  
    \end{itemize}
    Every generated image is mapped back to the canonical frame, creating a set of $K$ candidate predictions, the most-voted grids form the two attempts pass@1 and pass@2. 
\end{itemize}

\paragraph{Test-Time Noise Injection (Maze, Sudoku).}
We additionally inject noise during exploration, parametrized by $\Theta_{\text{noise}} = (r_{\text{noise}}, p_t, p_s, \sigma)$, where:
\begin{itemize}[leftmargin=12pt]
    \item[--] $r_{\text{noise}}$ is the active noise duration (typically the initial fraction of the rollout), after which noise is disabled;
    \item[--] $p_t$ and $p_s$ are the temporal and spatial Bernoulli probabilities of noise injection;
    \item[--] $\sigma$ is the standard deviation of the injected Gaussian noise $\mathcal{N}(0, \sigma^2)$.
\end{itemize}

\paragraph{Final Selection.}
Given $K$ candidate solutions $\{\hat{Y}_k\}_{k=1}^K$, we use these aggregation strategies:
\begin{itemize}[leftmargin=12pt]
    \item[--] \textbf{Conf@K (Maze-Hard, Sudoku):} Without access to ground truth $Y$, the model autonomously selects the candidate with the highest board confidence:
    \begin{equation*}
        k^* = \arg\max_{k \in \{1, \dots, K\}} c\left(X_k^{(D)}\right), \text{Conf@K} = \mathbb{I}\left(\hat{Y}_{k^*} = Y\right).
    \end{equation*}
    \item[--] \textbf{Pass@M via Majority Voting (ARC):} From the $K$ parallel rollouts, we select the $M$ most frequently predicted unique board configurations $\mathcal{S}_M = \{\hat{Y}_{(1)}, \dots, \hat{Y}_{(M)}\}$. The prediction is considered successful if any of the $M$ submissions matches the ground truth:
    \begin{equation*}
        \text{Pass@M} = \mathbb{I}\left(\exists \hat{Y} \in \mathcal{S}_M : \hat{Y} = Y^*\right).
    \end{equation*}
\end{itemize}

\subsection{FLOPs Estimation}
\label{subsec_flops_estimation}

\paragraph{Method.}
The floating-point operations (FLOPs) reported in the main paper correspond to inference on a \textit{single} board of size $S$, accounting the recurrence depth $D$ and the number of parallel trials $K$ used for test-time scaling. 

For PyTorch-based baselines (\textbf{DeepThink}, \textbf{AKOrN}, \textbf{(P)TRM}, \textbf{HRM}, and \textbf{LoopViT}), models were instantiated directly from their official codebase repositories using published configurations and evaluation settings ($S, D, K$). Single-step FLOPs were measured using PyTorch's \texttt{FlopCounterMode}. For all \textbf{NCA} models, implemented in JAX with configurations detailed in section~\ref{sec_experimental_settings}, single-step FLOPs were extracted from the compiled computation graph using the XLA cost analysis interface on CPU.

The total test-time compute is then calculated as $\text{FLOPs}_{\text{total}} = \text{FLOPs}_{\text{step}} \times D \times K$. 

All baseline configurations, model instanciations, parameter counts, and FLOPS estimation scripts are documented in the accompanying notebook \texttt{flops\_baselines.ipynb}.

\begin{table}
\centering
\setlength{\tabcolsep}{2pt} 
\footnotesize
\caption{TRM FLOPs per supervision step.}
\label{table_trm_flops_validation}
\begin{tabular}{lcc}
\toprule
\textbf{Method} & \shortstack[c]{\textbf{Maze-Hard / ARC} \\ \scriptsize(Att, $S{=}30{\times}30$)} & \shortstack[c]{\textbf{Sudoku-Extreme} \\ \scriptsize(MLP, $S{=}9{\times}9$)} \\
\midrule
PyTorch (FC) & 238.9 B & 25.6 B \\
PyTorch (Prof) & 187.5 B & 25.7 B \\
JAX (CPU) & 245.2 B & 26.2 B \\
JAX (A100 GPU) & 243.5 B & 25.8 B \\
\midrule
Relative ($\Delta_{\text{FC, CPU}}$) & 0.026 & 0.023 \\
\bottomrule
\end{tabular}
\end{table}

\paragraph{Validation.}
FLOP estimates can vary depending on framework-level operator definitions and the target accelerator backend. To quantify these variations, we reimplemented the TRM architecture~\citep{Jolicoeur-Martineau2025} in pure JAX and benchmarked the supervision-step FLOPs across four profiling configurations (Table~\ref{table_trm_flops_validation}).

In Table~\ref{table_results_across_benchmarks}, we report \textbf{PyTorch \texttt{FlopCounterMode}} for baselines (to capture all operations and activations without kernel-level omissions) and \textbf{JAX compiled on CPU} for Reasoning NCA (to evaluate exact mathematical operations without GPU/TPU hardware tile padding or compiler rewrites). As shown in Table~\ref{table_trm_flops_validation}, both chosen estimators align consistently with less than $3\%$ relative difference ($\Delta_{\text{FC, CPU}} < 3\%$), with JAX reporting slightly higher counts.

\paragraph{Limitations.}
While FLOPs offer a hardware-agnostic proxy for computational complexity, some limitations must be acknowledged. First, our estimation is \textbf{implementation-dependent}. For instance, our NCA implementation computes updates for all $H \times W$ cells before applying a binary mask $m_{i,j} \sim \operatorname{Bernoulli}(p_{\text{fire}})$ to emulate asynchronous updates. For $p_{\text{fire}} = 0.8$, this dense execution incurs a $1.25\times$ arithmetic overhead, resulting in higher reported FLOPs than an implementation where dormant cells perform zero operations. Second, FLOP counts measure raw arithmetic operations but do not account for \textbf{memory access patterns}, \textbf{parallelizability}, or \textbf{energy footprint}. Despite being hardware-dependent, we believe that such metrics could provide better insights into the scalability and energy advantages on current and/or future distributed hardware that FLOP counts alone do not capture.

\subsection{TTS Pruning Method}
\label{subsec_tts_pruning_details}

When scaling test-time compute by running an ensemble of $K$ parallel stochastic rollouts on a single puzzle, independent instances often converge to identical intermediate trajectories. Running duplicate trajectories to completion incurs substantial compute overhead without expanding exploratory breadth. To eliminate this redundancy, we introduce \textit{Niche-Capped Diversity Pruning}, an algorithm that periodically identifies congruent solution branches, caps the number of duplicate instances, and progressively concentrates compute on diverse hypotheses.

\paragraph{Algorithm Formulation.}
Consider an initial ensemble of $K$ parallel rollout trajectories $\mathcal{E}^{(0)} = \{X_1^{(0)}, \dots, X_K^{(0)}\}$ initialized with Gaussian noise $\sigma_{\text{init}}$ and rolled out for a total horizon of $D$ steps. We partition the trajectory into $m$ equidistant checkpoints spaced by stride $\Delta t = D / m$, corresponding to evaluation intervals $t \in \{\Delta t, 2\Delta t, \dots, (m-1)\Delta t\}$.

At each checkpoint $t$, pruning proceeds in four steps:
\begin{enumerate}
    \item[--] \textbf{Discrete State Extraction:} For each active trajectory $s \in \mathcal{E}^{(t)}$, the current board configuration $\hat{Y}_s^{(t)} \in \{0, \dots, V-1\}^{H \times W}$ is decoded from the cell readout slices:
    \begin{equation*}
        \hat{Y}_{s, i, j}^{(t)} = \arg\max_v \ell_{s, i, j}^{(t)}(v)
    \end{equation*}
    where $\ell_{s, i, j}^{(t)}(v)$ is the cosine similarity score between the cell state $x_{s, i, j, \text{out}}^{(t)}$ and token embedding $e_v$ (Section~\ref{subsec_architecture_details}).
    
    \item[--] \textbf{Niche Partitioning:} Trajectories are clustered into discrete equivalence classes or \textit{niches} $\mathcal{N}_c$ sharing identical predicted board states:
    \begin{equation*}
        \mathcal{N}_c = \left\{ s \in \mathcal{E}^{(t)} \;\middle|\; \hat{Y}_s^{(t)} = Y_c \right\}
    \end{equation*}
    where $Y_c$ denotes a unique candidate board configuration.

    \item[--] \textbf{Niche-Cap Filtering:} To prevent over-representation of any single attractor basin, each niche $\mathcal{N}_c$ is capped at a maximum capacity of $n_{\text{cap}}$ seeds (we use $n_{\text{cap}} = 3$). For niches with $|\mathcal{N}_c| > n_{\text{cap}}$, we retain the $n_{\text{cap}}$ instances with the highest global board confidence $c(X_s^{(t)})$ (or uniform sampling) and discard the remainder:
    \begin{equation*}
        \tilde{\mathcal{N}}_c = \operatorname{Top-}n_{\text{cap}}\left(\mathcal{N}_c, \; \text{by } c(X_s^{(t)})\right).
    \end{equation*}

    \item[--] \textbf{Ensemble Halving:} The total active ensemble is halved to target size $\lfloor K / 2 \rfloor$ by pooling surviving seeds from the capped niches, prioritizing representation across distinct niches to maximize hypothesis diversity before resuming rollouts.
\end{enumerate}

\paragraph{Exploration Noise Schedule.}
As in unpruned exploration (Section~\ref{subsec_testing_details}), test-time Gaussian perturbations, $\mathcal{N}(0, \sigma^2)$, are injected during the initial quarter of the trajectory ($r_{\text{noise}} = 0.25$) with temporal probability $p_t = 0.2$, spatial probability $p_s = 0.8$, and standard deviation $\sigma = 0.1$. This ensures rollouts explore divergent state-space regions before the first pruning interval culls duplicate attractors.

\paragraph{Theoretical FLOPs Reduction.}
For an initial ensemble $K$ unrolled over $D$ steps and halved at $m \in \{2, 4, 8\}$ equidistant intervals, total compute scales as:
\begin{equation*}
\begin{aligned}
    \text{FLOPs}(m) &= \sum_{i=0}^{m-1} \left( \frac{K}{2^i} \cdot \frac{D}{m} \right) \\
    &= \frac{2 - 2^{-(m-1)}}{m} \cdot (K \cdot D).
\end{aligned}
\end{equation*}
The relative FLOP savings are $1 - \frac{2 - 2^{-(m-1)}}{m}$, yielding exact savings of \textbf{25.0\%} for $m=2$, \textbf{53.1\%} for $m=4$, and \textbf{75.1\%} for $m=8$, with minimal effect on final accuracy across 110,000 Sudoku-Extreme boards.

\section{Experimental Settings}
\label{sec_experimental_settings}

\subsection{Maze}
\label{subsec_maze_settings}

Detailed architecture and training configurations are listed in Table~\ref{table_maze_sudoku_hparams}. Training took approximately 3 minutes for Maze-OOD and 2h30 for Maze-Hard (TPU v5lite).

At test time, states are initialized randomly ($\sigma_{\text{init}}$ as in training). 
For Maze-Hard, we perform parallel rollouts with noise injection for test-time scaling experiments. We use the same noise regime that in training ($p_{t}=0.1, p_{s}=0.3$) but vary noise magnitude $\sigma$ between 0 and 1 (as reported in Figure~\ref{figure_noise_generalization})  and apply it for the first quarter of the rollout ($r_{\texttt{noise}}=0.25$). 

\subsection{Sudoku}
\label{subsec_sudoku_settings}

Detailed architecture and training configurations are listed in Table~\ref{table_maze_sudoku_hparams}. Training took approximately 3h for Sudoku-OOD (TPU v5lite) and 22h for Sudoku-Extreme (TPU7x).

At test time, states are initialized randomly ($\sigma_{\text{init}}$ as in training). We perform parallel rollouts with noise injection for test-time scaling experiments. We use the same noise regime that in training ($p_{t}=0.1, p_{s}=0.2$) but vary noise magnitude $\sigma$ between 0 and 1 (as reported in Figure~\ref{figure_noise_generalization})  and apply it for the first quarter of the rollout ($r_{\texttt{noise}}=0.25$). 

\begin{table*}[h!]
\centering
\setlength{\tabcolsep}{4pt}
\footnotesize
\caption{Training hyperparameters for the Maze and Sudoku benchmarks.}
\label{table_maze_sudoku_hparams}
\begin{tabular}{lcccc}
\toprule
\textbf{Hyperparameter} & \textbf{Maze-OOD} & \textbf{Maze-Hard} & \textbf{Sudoku-OOD} & \textbf{Sudoku-Extreme} \\
\midrule
Grid size ($H \times W$) & $9 \times 9$ & $30 \times 30$ & $9 \times 9$ & $9 \times 9$ \\
Vocabulary size ($V$) & 4 & 5 & 10 & 10 \\
Channels ($C$) & 16 & 32 & 128 & 256 \\
I/O emb dim ($C_{\text{in}} / C_{\text{out}}$) & 16 / 2 & 32 / 3 & 32 / 32 & 32 / 32 \\
Sensing module & 2D conv & 2D conv & Fixed Attn & Fixed Attn \\
Sensing heads ($K_{\text{heads}}$) & 4 & 4 & 16 & 16 \\
Boundary padding & wall & wall & wrap & wrap \\
Update expansion ($E$) & 2 & 2 & 4 & 4 \\
Normalization ($U$) & None & None & GroupNorm (4) & GroupNorm (8) \\
State initialization $(\sigma_{\text{init}})$ & 0.15 & 0.25 & 0.15 & 0.15 \\
Cell fire rate ($p_{\text{fire}}$) & 0.8 & 0.8 & 0.8 & 0.8 \\
\midrule
Training steps & 5,000 & 20,000 & 150,000 & 200,000 \\
Batch size ($B$) & 64 & 32 & 256 & 800 \\
Pool size ($M / B$) & 4 & 2 & 4 & 4 \\
Batch seed ratio ($r_{\text{seed}}$) & 0.25 & 0.50 & 0.50 & 0.25 \\
Rollout length ($N$) & 100 & 256 & 48 & 96 \\
Chunk size ($N_{\text{chunk}}$) & 100 & 128 & 48 & 96 \\
Loss application & all chunk & all chunk & all chunk & last step \\
\midrule
Optimizer & AdamW & AdamW & AdamW & AdamW \\
Gradient clipping & 1.0 & 1.0 & 1.0 & 1.0 \\
Peak learning rate & $4 \times 10^{-4}$ & $5 \times 10^{-4}$ & $1 \times 10^{-4}$ & $1 \times 10^{-4}$ \\
Learning rate schedule & constant & cosine decay & cosine decay & cosine decay \\
Warmup ratio & 0 & 25\% & 2\% & 2\% \\
Weight decay & 0 & $1 \times 10^{-4}$ & $1 \times 10^{-4}$ & $1 \times 10^{-4}$ \\
EMA decay rate & 0.999 & 0.999 & 0.999 & 0.999 \\
Noise ($p_{t}, p_{s}, \sigma$) & $0.1, 0.2, 0.15$ & $0.1, 0.3, 0.15$ & $0.1, 0.2, 0.15$ & $0.1, 0.2, 0.15$ \\
Damage ($p_{\text{damage}}$) & 0.1 & 0.1 & 0.1 & 0.1 \\
Target swap ($p_{\text{swap}}$) & 0.1 & 0.1 & 0.1 & 0.1 \\
Data augmentation & 8 dihedral & None & permutations & permutations \\
\bottomrule
\end{tabular}
\end{table*}

\subsection{ARC-AGI-1}
\label{subsec_arc_settings}
Detailed architecture and training configurations are listed in Tables~\ref{table_arc_arch_hparams} and~\ref{table_arc_train_hparams}. Training took approximately 24h for ARC-AGI-1 (TPU 7x). 

At test time, we use $A=51$ offline augmentations and $R=64$ online augmentations per task of the public evaluation set ($K$=3264, as reported in Figure~\ref{figure_arc_ttt}). 

\begin{table*}[h!]
\centering
\setlength{\tabcolsep}{2pt}
\footnotesize
\begin{minipage}[t]{0.42\textwidth}
\centering
\caption{NCA architecture hyperparameters for ARC-AGI-1.}
\label{table_arc_arch_hparams}
\begin{tabular}{lc}
\toprule
\textbf{Hyperparameter} & \textbf{Value} \\
\midrule
Grid size ($H \times W$) & $32 \times 32$ \\
Vocabulary size ($V$) & 12  \\
Channels ($C$) & 512 \\
I/O emb dim ($C_{\text{in}} / C_{\text{out}}$) & 32 / 32 \\
Task emb dim ($C_{\text{task}}$) & 128  \\
Hidden state dim ($C_{\text{hid}}$) & 320 \\
Sensing module & Fixed Attn \\
Sensing heads ($K_{\text{heads}}$) & 32 \\
Boundary padding & zero \\
Update block & SwiGLU \\
Normalization ($U$) & GroupNorm (32)  \\
State initialization ($\sigma_{\text{init}}$) & 0.10 \\
Cell fire rate ($p_{\text{fire}}$) & 0.9 \\
\bottomrule
\end{tabular}
\end{minipage}
\hfill
\begin{minipage}[t]{0.55\textwidth}
\centering
\caption{Pre-training and TTT optimization hyperparameters for ARC-AGI-1.}
\label{table_arc_train_hparams}
\begin{tabular}{lcc}
\toprule
\textbf{Hyperparameter} & \textbf{Pre-training} & \textbf{TTT} \\
\midrule
Rollout length ($N$) & 64 & 64 \\
Loss window ($w$) & 16 & 16 \\
Training steps & 650,000 & 10,000 / task \\
Batch size ($B$) & 128 & 8 \\
\midrule
Optimizer & AdamW & Adam \\
Gradient clipping & 1.0 & 1.0 \\
Peak learning rate & $1 \times 10^{-3}$ & $1 \times 10^{-4}$ \\
LR schedule & Cosine ($\alpha=0.1$) & Cosine ($\alpha=0.0$) \\
Weight decay & $1 \times 10^{-4}$ & 0 \\
Data augmentation & Online+Re-ARC & Online+Offline \\
\bottomrule
\end{tabular}
\end{minipage}
\end{table*}

\subsection{Visual Sudoku}
\label{subsec_visual_sudoku_settings}

Detailed architecture and training configurations are listed in Table~\ref{table_visual_sudoku_hparams}. The architecture and approach served to simply validate that an NCA model could in principle solve a large, complex task like Visual Sudoku. This was done by taking an \enquote{off-the-shelf} NCA and associated training pipeline, and directly applying to the Visual Sudoku image-to-image task. As a result, the training is not optimised for this task, and incorporates many of the assumptions made for the small model, such as back-propagation through full unrolls of the entire trajectory of 1024 steps. This results in a compute intensive training regime which we believe could be significantly optimised. Training took approximately three days on 64 v5p TPUs machines. The training also occasionally suffered from instabilities, where the loss would spike. When this happened, training was resumed from a previous, clean, checkpoint, with a lower learning rate, to move past the spike, then resumed with the original learning rate.

\begin{table}[h!]
\centering
\setlength{\tabcolsep}{6pt}
\footnotesize
\caption{Training hyperparameters for Visual Sudoku.}
\label{table_visual_sudoku_hparams}
\begin{tabular}{lc}
\toprule
\textbf{Hyperparameter} & \textbf{Visual Sudoku} \\
\midrule
Grid size ($H \times W$) & $256 \times 256$ (4 px pad) \\
Channels ($C$) & 256 \\
Visual channels ($C_{\text{in}} / C_{\text{out}}$) & 1 / 1 \\
Sensing module & 2D Conv (Sobel + Identity) \\
Sensing heads ($K_{\text{heads}}$) & 3 (fixed) \\
Boundary padding & wrap \\
Update expansion ($E$) & 8 \\
Normalization & None \\
State initialization & $\mathcal{U}(-0.1, 0.1)$ \\
Cell fire rate ($p_{\text{fire}}$) & 0.5 \\
\midrule
Training steps & 93,000 \\
Rollout length ($N$) & 1024 \\
Loss application & All steps \\
\midrule
Optimizer & AdamW \\
Gradient clipping & 1.0 \\
Peak learning rate & $1 \times 10^{-4}$ \\
Weight decay & $1 \times 10^{-4}$ \\
EMA decay rate & 0.999 \\
\bottomrule
\end{tabular}
\end{table}

\section{Additional Results}
\label{sec_additional_results}

\subsection{Fixed Attention Analysis}
\label{subsec_fixed_attention}

\begin{figure*}
  \centering
  \includegraphics[width=\textwidth]{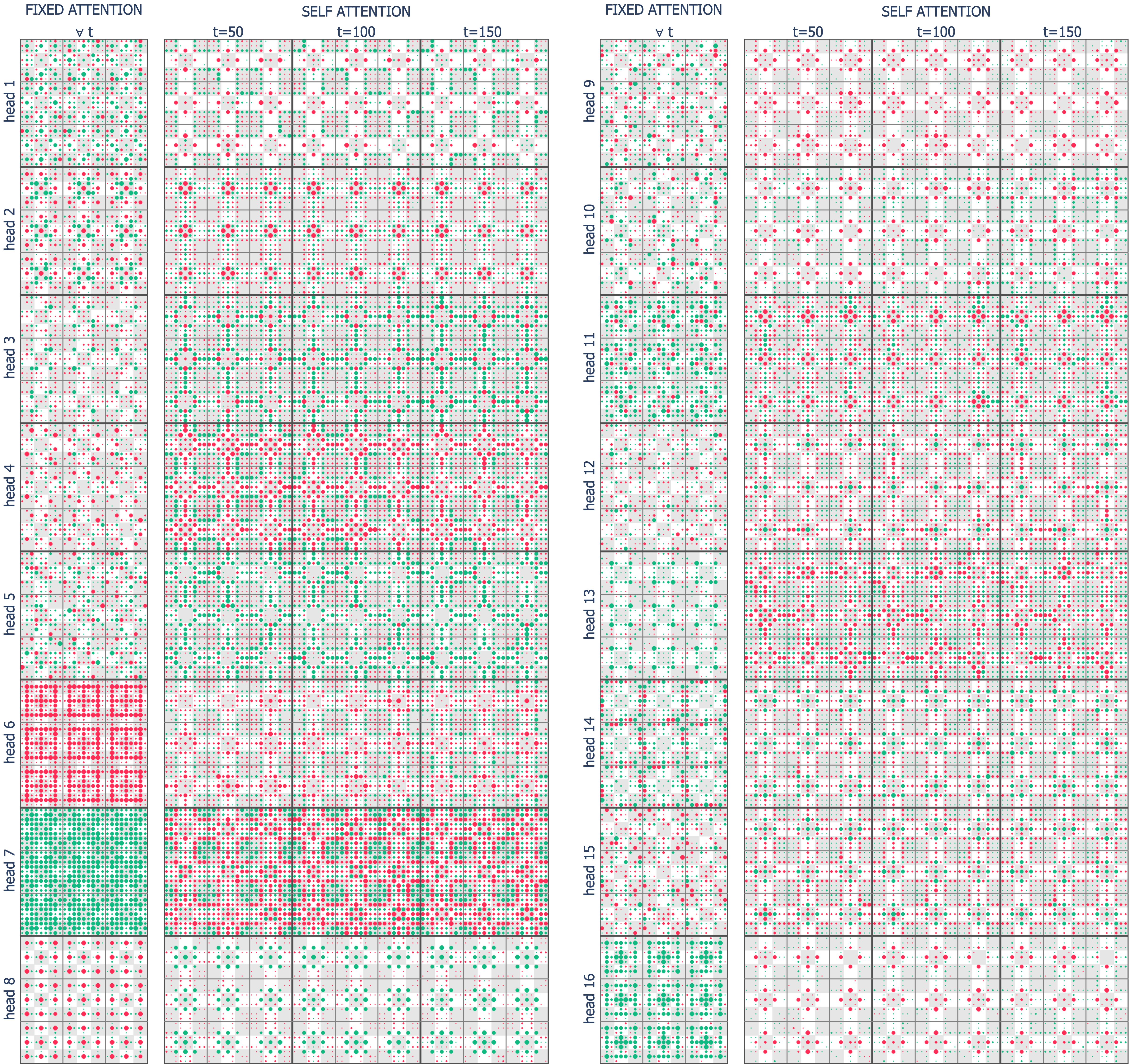}
  \caption{\textbf{Self-attention converges to static spatial patterns.} 
Attention weights across the 16 perception heads for fixed attention (left) and self-attention at rollout steps $t \in \{50, 100, 150\}$ (right). Each panel shows the $9 \times 9$ Sudoku grid, with each cell displaying 9 dots showing attention weight to its $3 \times 3$ Moore neighborhood (size indicates weight magnitude; green is positive, pink is negative). Self-attention weights, derived from query-key dot product at rollout step $t$, have converged to static spatial patterns that do not depend on cell states nor time step. Fixed attention parameterizes this inter-cell coupling directly, leading to more efficient learning.}
  \label{figure_sa_convergence}
\end{figure*}

We tested an attention-based perception module to provide cells with a more expressive, position-dependent, and data-dependent way to attend to their neighbors compared to a convolution perception module. We first implemented a traditional self-attention perception module with key-query dot products within the $3\times3$ neighborhood to obtain attention scores. Yet, inspecting the learned weights revealed that self-attention converged to static, spatially symmetric patterns independent of time step (Figure~\ref{figure_sa_convergence}); rather than routing information dynamically based on cell states. 

\begin{figure}[h!]
  \centering
  \includegraphics[width=0.5\textwidth]{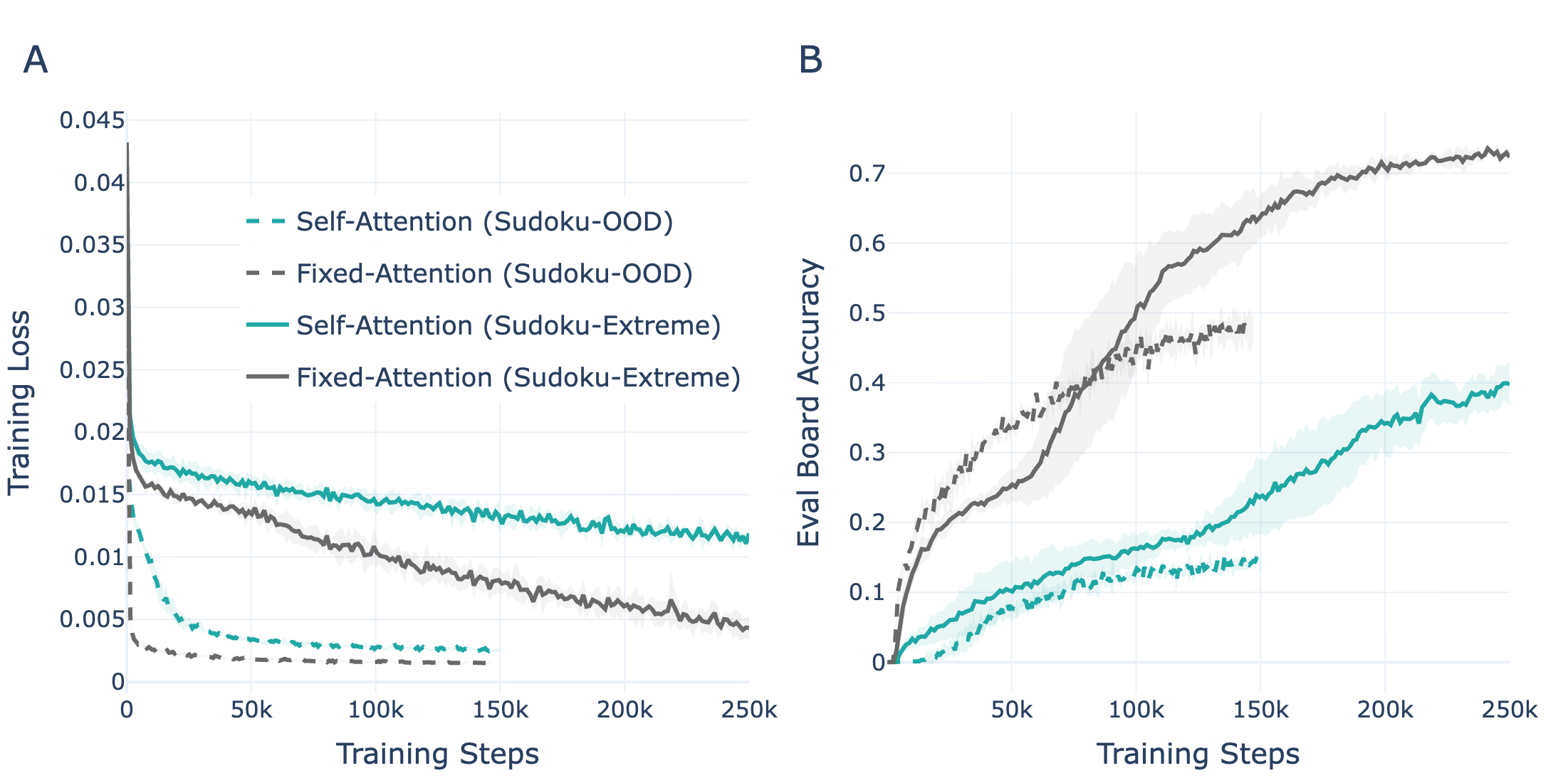}
  \caption{\textbf{Fixed-attention ablation study.} 
  Training loss (\textbf{A}) and board accuracy (\textbf{B}) show that fixed attention consistently trains faster and achieves higher solve rates than the standard self-attention on Sudoku-OOD and Sudoku-Extreme.
  Mean-std curves over 3 training seeds are shown.
  }
  \label{figure_sa_vs_fa}
\end{figure}

Motivated by this observation, we replaced dynamic self-attention with \enquote{fixed attention}, directly parameterizing static, position-dependent mixing weights across the $3 \times 3$ neighborhood. This change eliminates the need to compute query-key dot products at every rollout step. Beyond computational savings, we found that directly parameterizing this fixed attention significantly improved training, achieving lower training loss and higher final board accuracy across both Sudoku-OOD and Sudoku-Extreme benchmarks (Figure~\ref{figure_sa_vs_fa}). 

\subsection{Impact of the Locality Bottleneck}
\label{subsec_locality_bottleneck}

To study how spatial locality shapes iterative reasoning dynamics, we compare NCAs with TRM \citep{Jolicoeur-Martineau2025}. Unlike NCAs, TRM uses all-to-all connectivity and synchronous updates, allowing every grid token to attend to all other tokens via global self-attention at each step.

We evaluate both models on the 1000 test instances of Maze-Hard ($30 \times 30$) over $t = 1, \dots, D$ steps, where $t$ denotes each time step at which the output prediction is updated. In TRM, the internal state consists of two components: a high-level answer state $y$ from which predictions are decoded (analogous to our $C_{\text{out}}$ channels) and a low-level latent reasoning state $z$ (analogous to our $C_{\text{hid}}$ channels). The answer state $y$ is updated once every $L_{\text{cycles}} = 4$ updates of the latent state $z$. TRM is evaluated with $N_{\text{sup}} = 16$ supervision steps and $H_{\text{cycles}} = 3$ prediction updates per supervision step, resulting in a maximum rollout depth of $D = N_{\text{sup}} \times H_{\text{cycles}} = 48$ prediction steps. For the NCA, we use $D = 200$ steps. As TRM did not release official checkpoints, we use the reproduction by \url{https://github.com/gaoxin492/TinyRecursiveModels}.

\begin{table}[h!]
  \centering
  \setlength{\tabcolsep}{2pt}
  \caption{\textbf{The locality bottleneck enforces iterative reasoning.} 
  Time-to-Solve (mean $\pm$ std) on the mutually solved subset of Maze-Hard.}
  \label{table_locality_bottleneck}
  \small
  \begin{tabular}{lcc}
    \hline
    Model & Update Steps ($D$) & Time-to-Solve ($t_{\text{solve}}$) \\
    \hline
    TRM & 48  & 3.3 $\pm$ 2.3 \\
    NCA & 200 & 42.3 $\pm$ 21.1 \\
    \hline
  \end{tabular}
\end{table}

We define $t_{\text{solve}}$ as the earliest step at which the decoded prediction matches the exact ground-truth ($A^*$) path and remains stably correct for all remaining steps through the end of the rollout. Within their respective budgets, both models achieve similar solve rates: 787/1000 for TRM and 790/1000 for NCA. Table~\ref{table_locality_bottleneck} reports the average $t_{\text{solve}}$ computed over the subset of 679 mazes that both NCA and TRM solved. TRM converges in only $\sim 3$ prediction steps (corresponding to $\sim 12$ updates of the latent vector $z$), whereas the NCA requires $\sim 42$ steps.

\begin{figure}[h!]
  \centering
  \includegraphics[width=0.5\textwidth]{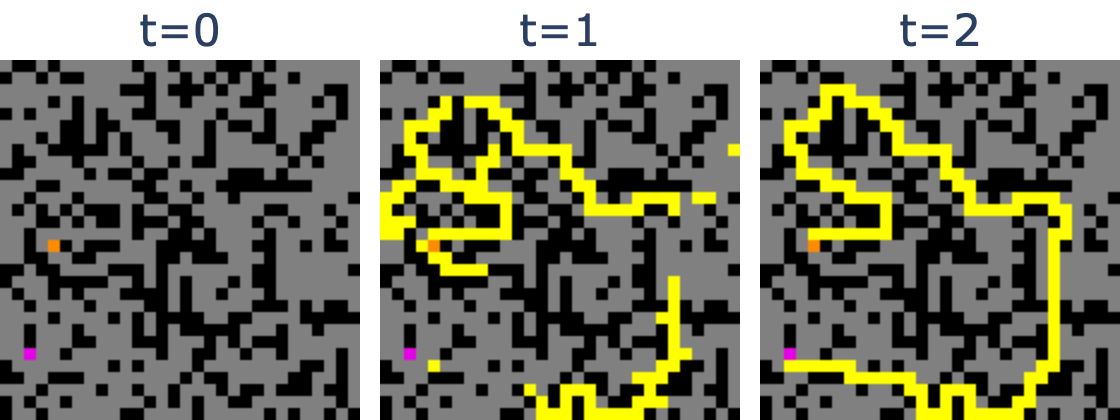}
  \caption{\textbf{TRM converges in very few steps.}}
  \label{figure_trm_solving}
\end{figure}

Visualizing the intermediate predictions on an example maze (Figure S\ref{figure_trm_solving}) shows that TRM \enquote{jumps} to the solution very quickly (in 2 prediction steps). In contrast, NCAs show contiguous, wave-like path exploration (Figure \ref{figure_reasoning_skills}). While NCAs require more steps for solving mazes due to this locality constraint, it makes the NCA's reasoning process quite interpretable with an observable spatial \enquote{chain-of-thought} that can be visually tracked on the grid.

\begin{figure*}[h!]
  \centering
  \includegraphics[width=\textwidth]{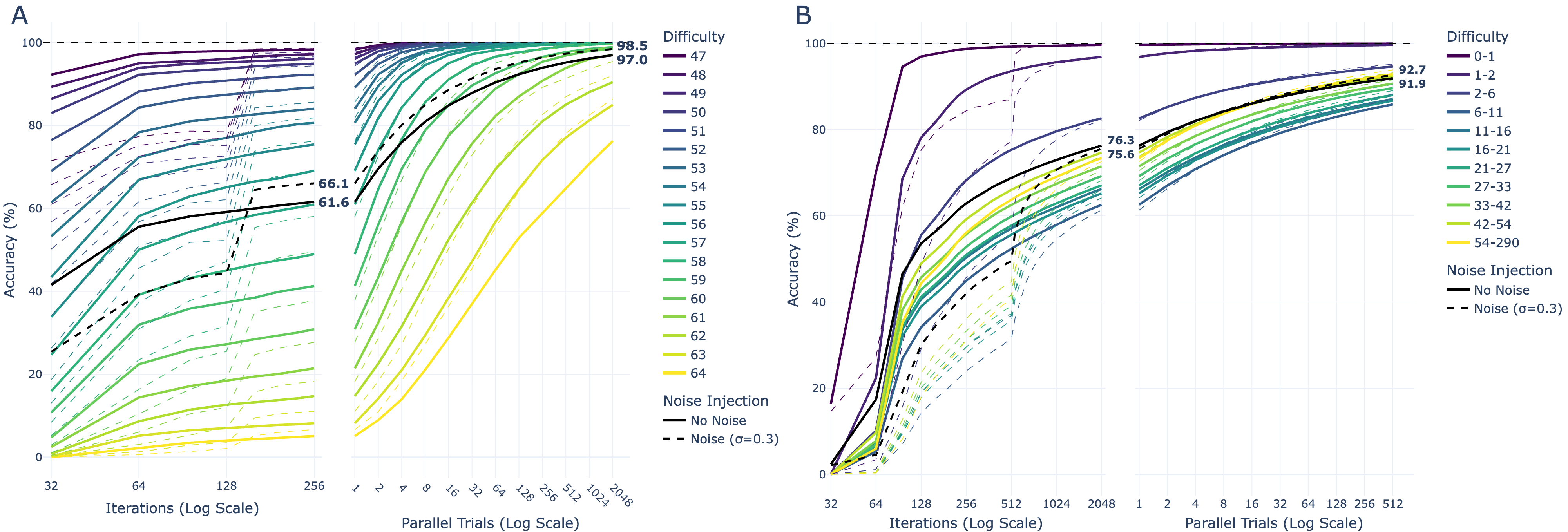}
  \caption{\textbf{Extended test-time scaling results on Sudoku benchmarks.} 
  Evaluation on \textbf{(A)} Sudoku-OOD and \textbf{(B)} Sudoku-Extreme across, rollout iterations (left sub-panel) and parallel trials (right sub-panel). Colored curves show test accuracy for different difficulties without noise (solid lines) and with noise injection (dashed lines). Metrics are averaged over 3 test seeds.}
  \label{figure_sudoku_full_tts}
\end{figure*}

\subsection{Extended Test-Time Scaling Results}
\label{subsec_tts_full_results}

\subsubsection{Sudoku}

We find that parallel trials significantly improve performance on Sudoku benchmarks (Figure~\ref{figure_sudoku_full_tts}). Sampling multiple stochastic trajectories increases the likelihood of finding the correct solution, and our confidence-based selection mechanism reliably identifies it among the $K$ candidates. On Sudoku-OOD, this strategy boosts accuracy from $\sim$50\% (K=1, D=48) to 97\% (K=2048, D=256); and on Sudoku-Extreme from $\sim$50\% (K=1, D=96) to 91.9\% (K=512, D=2048). 

We also observe that injecting noise at test-time is beneficial: while causing an initial accuracy drop when applied, adding noise at inference eventually slightly exceeds the performance of the no-noise variant by few percents rising accuracy to 98.5\% for Sudoku-OOD and 92.7\% for Sudoku-Extreme.

Interestingly, we find that the difficulty ordering in Sudoku-Extreme differs from empirical difficulty: NCAs struggle most with intermediate difficulty bins ($[6\text{--}11]$ and $[11\text{--}16]$ backtracks in \texttt{tdoku}) rather than the highest-backtrack bins ($[54\text{--}290]$). This suggests that intermediate boards contain more densely entangled candidate constraints across multiple cells, creating local minima that are harder for NCAs to resolve.

\begin{figure*}[h!]
  \centering
  \includegraphics[width=\textwidth]{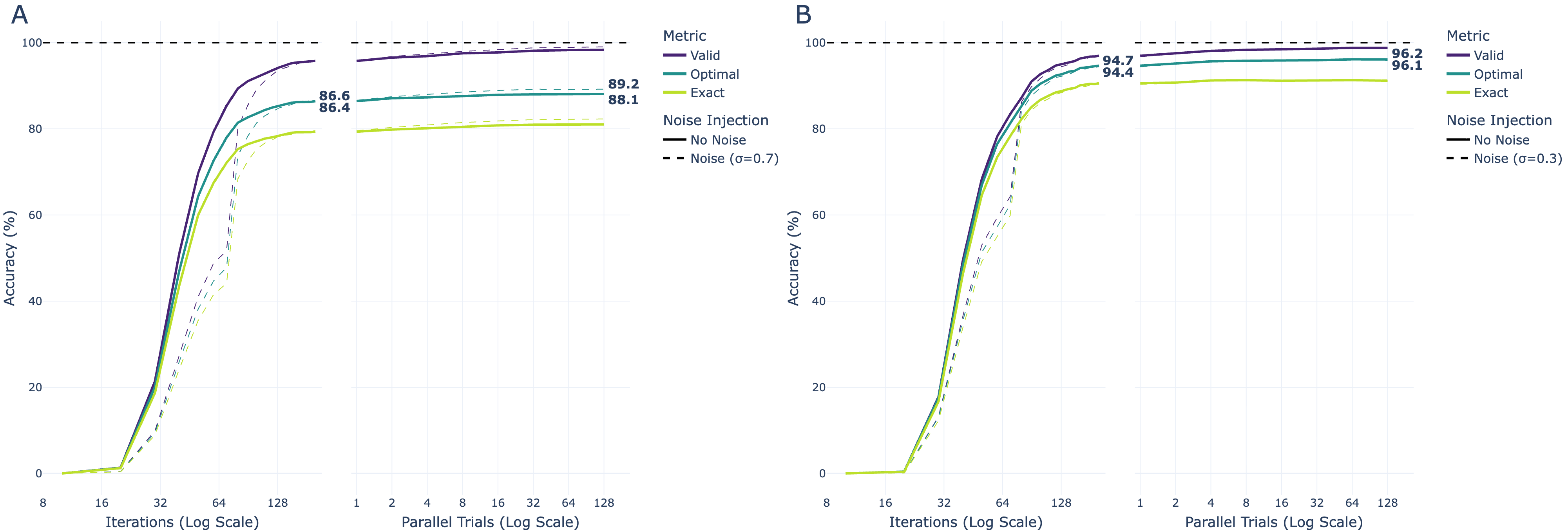}
  \caption{\textbf{Extended test-time scaling results on Maze benchmarks.} 
  Evaluation on Maze-Hard \textbf{(A)} with $A^*$ targets and \textbf{(B)} with BFS targets, across rollout iterations (left sub-panel) and parallel trials (right sub-panel). Colored curves show test accuracy for different difficulties without noise (solid lines) and with noise injection (dashed lines). Metrics are averaged over 3 test seeds. }
  \label{figure_maze_full_tts}
\end{figure*}

\subsubsection{Maze}

On Maze-Hard, test-time scaling is also beneficial but provides smaller gains, pushing results from $\sim$85\% (K=1, D=128) to 88.1\% (K=128, D=128) when trained on $A^*$ targets, and from $\sim$92\% (K=1, D=128) to 96.1\% (K=128, D=128) when trained on BFS targets; with more than 98\% of predictions being valid solutions in both cases. We see again small gains with noise-injection rising accuracy to 89.2\% on the model trained on $A^*$ targets. Overall, training on BFS targets in Maze-Hard instead of $A^*$ targets significantly improves performances.

\subsubsection{ARC-AGI-1}

\begin{figure*}[h!]
  \centering
  \includegraphics[width=\textwidth]{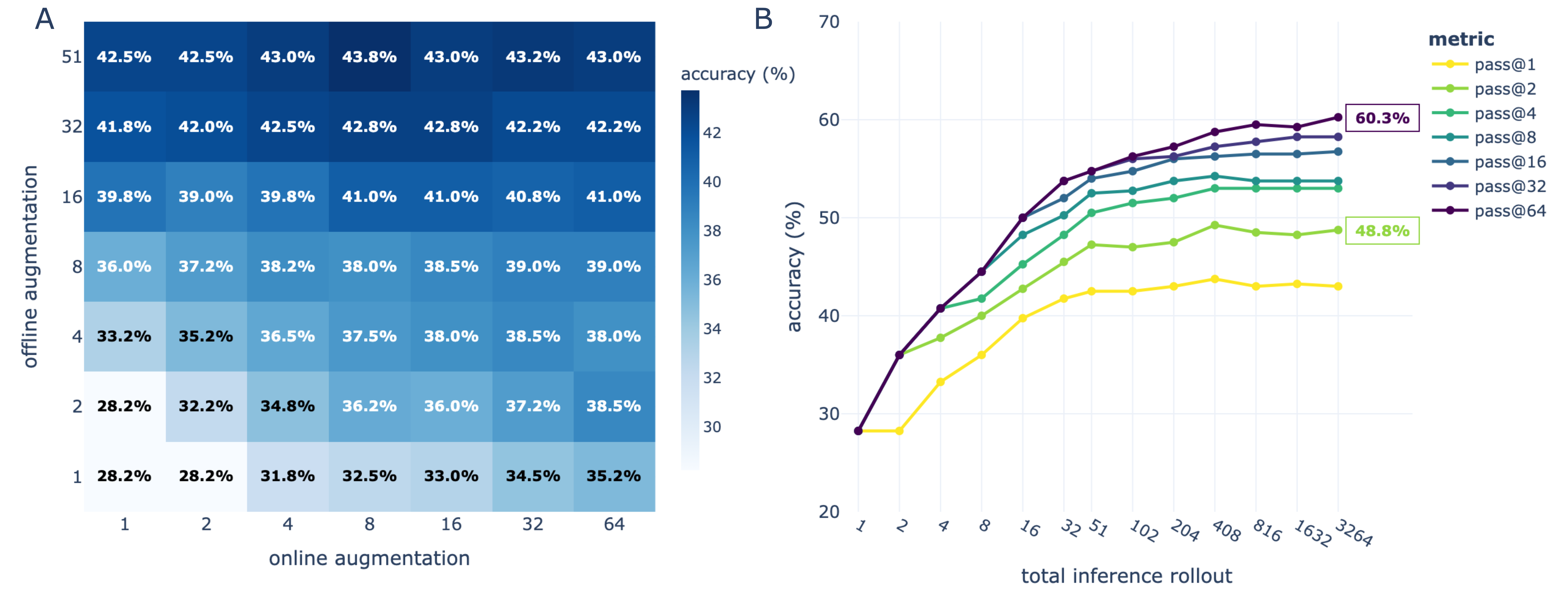}
    \caption{\textbf{Test-time compute scaling on ARC-AGI public evaluation set.} \textbf{(A)} Pass@1 majority-voting accuracy across increasing numbers of offline and online augmentations. \textbf{(B)} Pass@$K$ scaling under the offline-first-augmentation policy, reaching 60.3\% Pass@64. }
  \label{figure_arc_ttt_full}
\end{figure*}

In ARC-AGI-1, the use of parallel stochastic rollouts from different augmented inputs similarly increases pass@2 performances on the evaluation set from 28.25\% (K=1, D=64) to 48.75\% (K=3264, D=64). Moreover, considering Pass@64 pushes performance further to 60.3\%, suggesting that some of the stochastic rollouts successfully discover valid solutions, but the present candidate selection mechanism is unable to reliably select it. Using an NCA ensemble (NCA E) of 3 models independently fine-tuned further boosts performances to 63.0\% Pass@64.

\begin{figure}[h!]
  \centering
  \includegraphics[width=0.49\textwidth]{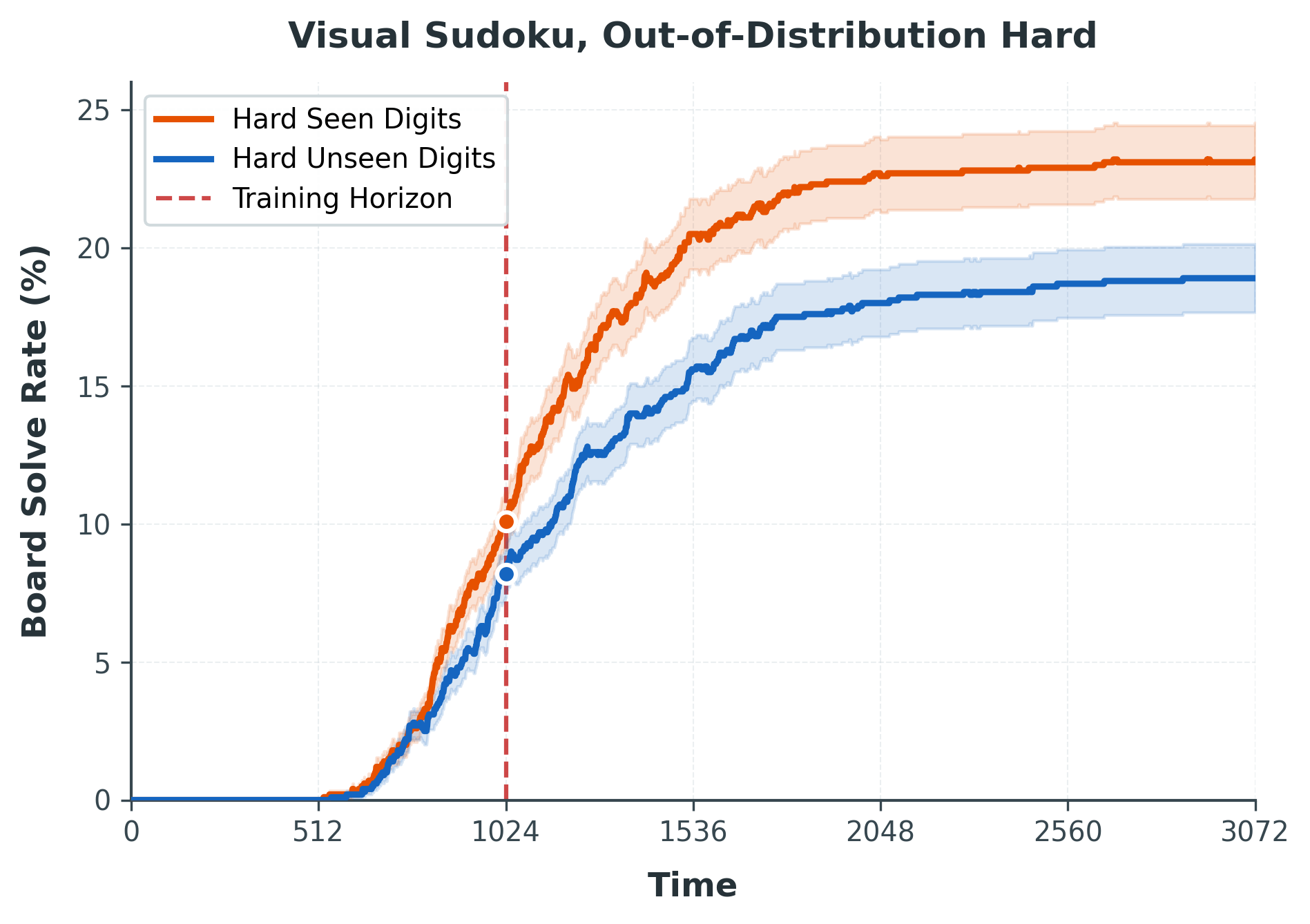}
  \caption{\textbf{Test-time scaling via extended rollouts on Visual Sudoku \textit{hard} puzzles}, rendered both with unseen and seen MNIST digits. The accuracy improves from approximately 10\% to 23\% with additional test time compute with seen, and from approximately 8\% to 19\% with unseen.}
  \label{figure_visual_tts}
\end{figure}

\subsubsection{Visual-Sudoku}
We observe that running the model for longer rollouts than used during training improves accuracy (Figure \ref{figure_visual_tts}).

\subsection{Extended TTS Pruning Results}
\label{subsec_tts_pruning_extended}

\begin{figure}
  \centering
  \includegraphics[width=0.5\textwidth]{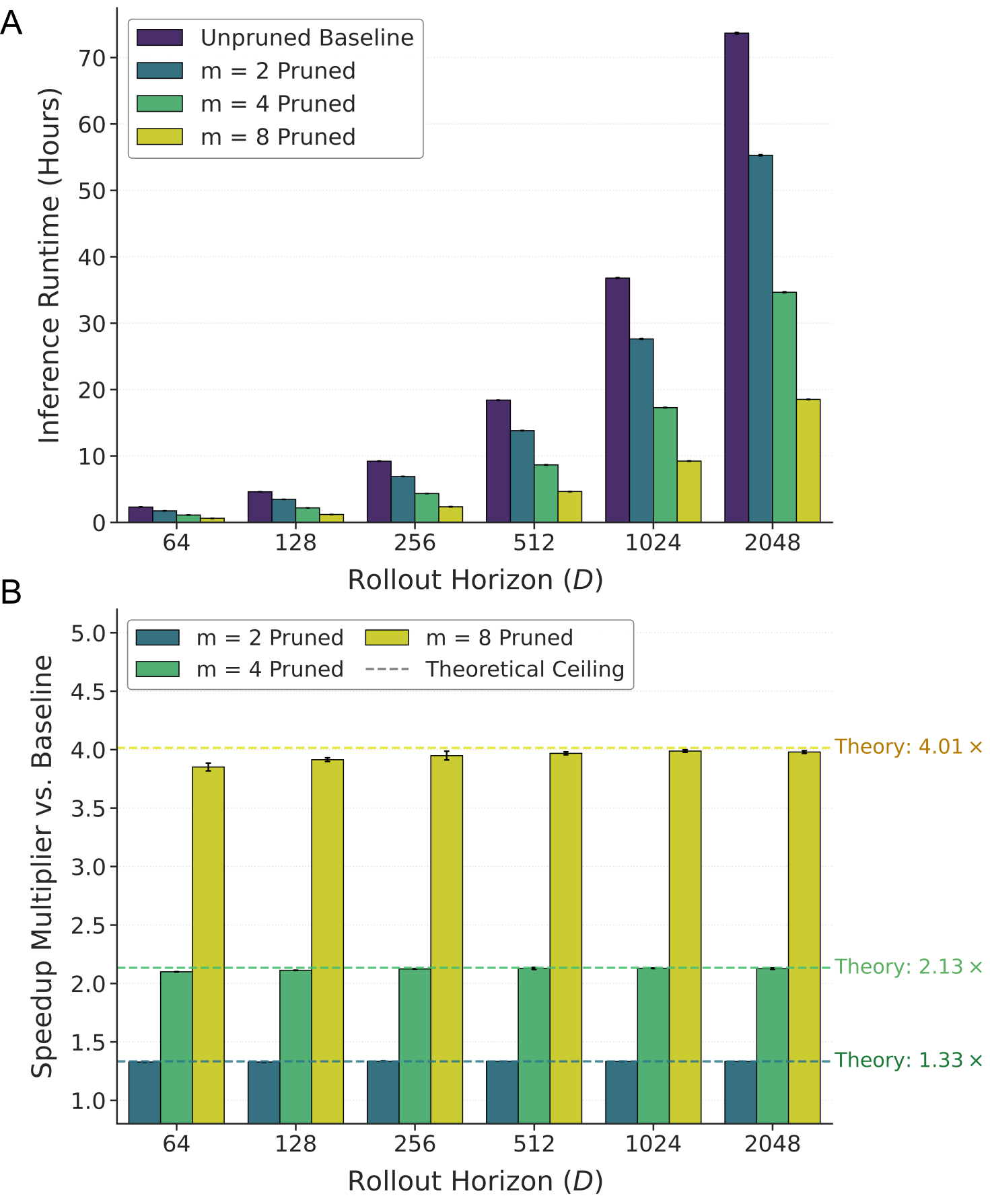}
  \caption{End-to-end wall-clock inference runtime and speedup of test-time pruning on Sudoku-Extreme. (A) Total inference runtime (hours on NVIDIA H100 GPUs across $110{,}000$ test puzzles) for the unpruned baseline ($K=512$) and pruning divisors $m \in \{2, 4, 8\}$ across rollout horizons $D \in \{64, \dots, 2048\}$ (error bars indicate $\pm 1$ SD over 3 seeds). (B) Empirical wall-clock speedup multiplier ($\text{Time}_{\text{base}} / \text{Time}_{\text{pruned}}$) versus rollout horizon $D$. Horizontal dashed lines indicate the theoretical FLOP-reduction ceilings $\frac{m}{2 - 2^{-(m-1)}}$ ($1.33\times$, $2.13\times$, and $4.01\times$ for $m=2, 4, 8$); measured speedups closely track the theoretical limits across all horizons ($<1\%$ sorting and compaction overhead).}
  \label{figure_tts_pruning_time}
\end{figure}

To evaluate the practical serving efficiency and hardware translation of niche-capped pruning, we measure the end-to-end wall-clock inference duration across all horizons $D\in [64,...,2048]$ on NVIDIA H100 GPUs (Figure~\ref{figure_tts_pruning_time}A). Unpruned baseline rollouts at peak horizon ($D=2048$) require $18.42\text{ hours}$ per shard (110,000 Sudoku-Extreme test puzzles). Progressive niche-capping reduces this duration dramatically to $13.82\text{ hours}$ ($m=2$), $8.66\text{ hours}$ ($m=4$), and $4.63\text{ hours}$ ($m=8$), cutting evaluation turnaround time and cloud serving costs by up to $13.79\text{ hours}$ ($3.98\times$). We also measure the empirical speedup ($\text{Time}_{\text{base}} / \text{Time}_{\text{pruned}}$) against the theoretical ceiling $\left[ \frac{m}{2 - 2^{-(m-1)}} \right]$. Across all evaluated rollout horizons ($D = 64 \dots 2048$), the measured acceleration closely tracks theoretical limits (Figure~\ref{figure_tts_pruning_time}B). This confirms that our vectorized JAX implementation incurs $<0.9\%$ sorting overhead, allowing theoretical FLOP formulations to serve as good predictors of real-world accelerator limits.

\end{document}